\documentclass[journal]{IEEEtai}

\usepackage[colorlinks,urlcolor=blue,linkcolor=blue,citecolor=blue]{hyperref}
\usepackage{color,array}
\usepackage{pifont}
\newcommand{\cmark}{\ding{51}}
\newcommand{\xmark}{\ding{55}}
\usepackage{graphicx}
\usepackage{amsmath}
\usepackage{lettrine}
\usepackage{multirow}
\usepackage{dirtytalk}
\usepackage{amsfonts}
\usepackage{float}
\usepackage{balance}
\usepackage[T1]{fontenc}
\usepackage[numbers]{natbib}

\usepackage[euler]{textgreek}

\usepackage[tableposition=top]{caption}

\begin{document}
\title{Challenges and Countermeasures for Adversarial Attacks on Deep Reinforcement Learning} 

\author{Inaam Ilahi\IEEEauthorrefmark{1}, Muhammad Usama\IEEEauthorrefmark{1}, Junaid Qadir, \IEEEmembership{Senior Member, IEEE}, Muhammad Umar Janjua, Ala Al-Fuqaha, \IEEEmembership{Senior Member, IEEE}, Dinh Thai Hoang, \IEEEmembership{Member, IEEE}, and Dusit Niyato, \IEEEmembership{Fellow, IEEE}
\thanks{\IEEEauthorrefmark{1}Equal contribution}
\thanks{Inaam Ilahi\IEEEauthorrefmark{2}, Muhammad Usama\IEEEauthorrefmark{3}, and Muhammad Umar Janjua\IEEEauthorrefmark{5} are with the Information Technology University (ITU), Lahore, Pakistan (e-mail: mscs18037@itu.edu.pk\IEEEauthorrefmark{2}, muhammad.usama@itu.edu.pk\IEEEauthorrefmark{3}, umar.janjua@itu.edu.pk\IEEEauthorrefmark{5}).}
\thanks{Junaid Qadir is with the Information Technology University (ITU, Lahore, Pakistan and Qatar University, Doha, Qatar. (email: junaid.qadir@itu.edu.pk).}
\thanks{Ala Al-Fuqaha is with the Hamad bin Khalifa University (HBKU), Doha, Qatar (email: aalfuqaha@hbku.edu.qa).}
\thanks{Dinh Thai Hoang is with the University of Technology Sydney (UTS), Australia (email: hoang.dinh@uts.edu.au).}
\thanks{Dusit Niyato is with the Nanyang Technological University (NTU), Singapore (email: dniyato@ntu.edu.sg).}
}
\markboth{Journal of IEEE Transactions on Artificial Intelligence, Vol. 00, No. 0, September 2021}
{Ilahi \MakeLowercase{\textit{et al.}}: Challenges and Countermeasures for Adversarial Attacks on Deep Reinforcement Learning}

\maketitle

\begin{abstract}
Deep Reinforcement Learning (DRL) has numerous applications in the real world thanks to its ability to achieve high performance in a range of environments with little manual oversight. Despite its great advantages, DRL is susceptible to adversarial attacks, which precludes its use in real-life critical systems and applications (e.g., smart grids, traffic controls, and autonomous vehicles) unless its vulnerabilities are addressed and mitigated. To address this problem, we provide a comprehensive survey that discusses emerging attacks on DRL-based systems and the potential countermeasures to defend against these attacks. We first review the fundamental background on DRL and present emerging adversarial attacks on machine learning techniques. We then investigate the vulnerabilities that an adversary can exploit to attack DRL along with state-of-the-art countermeasures to prevent such attacks. Finally, we highlight open issues and research challenges for developing solutions to deal with attacks on DRL-based intelligent systems.
\end{abstract}

\begin{IEEEImpStatement}
Deep Reinforcement Learning (DRL) has numerous real-life applications ranging from autonomous driving to healthcare. It has demonstrated superhuman performance in playing complex games like Go. However, in recent years, many researchers have identified various vulnerabilities of DRL. Keeping this critical aspect in mind, in this paper we present a comprehensive survey of different attacks on DRL and various countermeasures that can be used for robustifying DRL. To the best of our knowledge, this survey is the first attempt at classifying the attacks based on the different components of the DRL pipeline. This paper will provide a roadmap for the researchers and practitioners to develop robust DRL systems.
\end{IEEEImpStatement}


\section{Introduction}
\label{sec: intro}

The ultimate goal of research in Artificial Intelligence (AI) is to develop artificial general intelligence (AGI) agents that can perform similar activities as humans in a more efficient manner. This long-standing challenge for developing such agents is no longer a pipe dream thanks to rapid growth in computational AI and ML technologies. In the last decade, ML and especially Deep Learning (DL) has revolutionized fields like computer vision, language processing, etc. ML is divided into three categories \cite{abu2012learning}: namely, \textit{supervised, unsupervised,} and \textit{reinforcement learning}. In \textit{supervised learning}, training data along with the corresponding labels are available for decision making. Supervised learning is by far the most well-studied branch of ML for problems with labeled data, which has a lot of applications in practice such as object recognition, speech recognition, spam detection, pattern recognition, and many more. In \textit{unsupervised learning}, the target is to infer the underlying patterns and structures from unlabelled data. \textit{Reinforcement learning} (RL) is defined as a learning process that focuses on finding the best strategies for agents based on the interactions with the surrounding environment. Unlike supervised and unsupervised learning processes which need training data to learn, RL agents can learn in an online manner, based on observations obtained through real-time interactions with the environment.\par

RL utilizes a trial-and-error process to solve sequential decision-making problems in robotics, control, and many other real-world problems. RL algorithms also have some limitations to be utilized in practice mainly due to their slow learning process and inability to learn in complex environments. Recently, a new technique combining the advancement of DL, called deep reinforcement learning (DRL) has been introduced \cite{arulkumaran2017deep}. DRL has shown great results in many complex decision making processes such as designated task completion in robotics \cite{gu2017deep}, navigating driver-less autonomous vehicles \cite{peng2017deeploco}, \cite{fayjie2018driverless}, healthcare \cite{raghu2017continuous}, financial trading \cite{deng2016deep}, smart grids management \cite{franccois2016deep}, automated transportation management \cite{madu2017urban}, wireless and data networks management \cite{luong2018applications}, and for playing games such as Pong \cite{mnih2015human}, Go \cite{silver2016mastering}, etc. In 2017, DRL beat the human champions in the game of Go \cite{silver2017mastering} and most recently a team of five DRL agents has beaten the world champion human team in Dota2 matches \cite{berner2019dota}. This shows that DRL is promising and can address highly complex and time-sensitive decision-making problems in real-time.\par

\begin{figure*}[!t]
\centering
\includegraphics[trim=0cm 5cm 1.7cm 2cm,clip=true, width=1\textwidth]{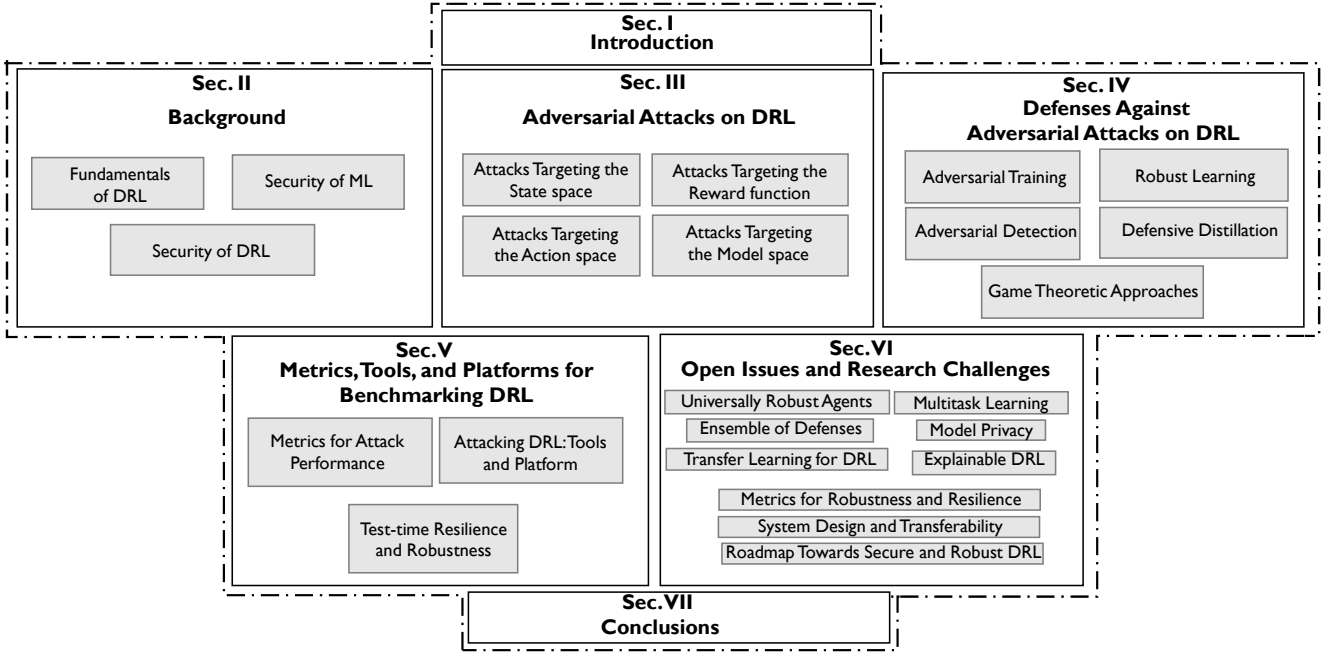}
\caption{Organization of the paper.}
\label{fig:outline}
\end{figure*}

With the rapid adoption of the DRL in critical real-world applications, the security of DRL has become a very important area of research \cite{kiran2020deep}, \cite{zhang2020deep}. Recently DRL is vulnerable to \textit{adversarial attacks}, where an imperceptible perturbation is added to the input to the DRL schemes with a pre-defined goal of causing a malfunction in the working of DRL \cite{behzadan2017vulnerability}. Thus, it is crucial to understand the types and nature of these vulnerabilities and their potential mitigation procedures before deploying DRL-based real-life critical systems (e.g. smart grids and autonomous vehicles). Here we want to note that the security of supervised and unsupervised ML is well studied in the literature \cite{akhtar2018threat}, but the security of the DRL has not yet received similar attention. In 2018, Behzadan and Munir \cite{behzadan2018faults} reviewed the security vulnerabilities and open challenges in DRL. Although it provides a decent initial review of the security concerns, it does not properly cover the security issues associated with four major components of the DRL pipeline (state, action, model, and reward) and related robustness mechanisms. Furthermore, the survey does not cover the recent studies. We aim to fulfill these requirements by providing a more comprehensive survey on attacks and defense techniques together with a discussion of the future research directions on DRL.

\textit{\textbf{Contributions of this paper:}} In this paper, we build upon the existing literature available on security vulnerabilities of DRL and their countermeasures and provide a comprehensive and extensive review of the related work. The major contributions of this paper are as follows:

\begin{itemize}
 \item We provide the DRL fundamentals along with a non-exhaustive taxonomy of advanced DRL algorithms. 
 \item We present a comprehensive survey of adversarial attacks on DRL and their potential countermeasures.
 \item We discuss the available benchmarks and metrics for the robustness of DRL.
 \item Finally, we highlight the open issues and research challenges in the robustness of DRL and introduce some potential research directions.
\end{itemize}

\textit{\textbf{Organization of the Paper}}:
The organization of this paper is depicted in Figure \ref{fig:outline}. An overview of the challenges faced by ML and DRL schemes has been provided in Section \ref{sec: background}. Section \ref{sec:attacks} presents a comprehensive review of adversarial ML attacks on the DRL pipeline. A detailed overview of countermeasures proposed in the literature to ensure robustness against adversarial attacks is presented in Section \ref{sec:adv_ML_def}. Section \ref{sec:bench} presents the available benchmarking tools and metrics along with open research problems in DRL. Section \ref{sec:open} describes the open issues and research challenges in designing adversarial attacks and robustness mechanisms for DRL. Finally, we conclude the paper in Section \ref{sec:con}. For the convenience of the reader, a summary of the salient acronyms used in this paper is presented in Table \ref{tab:acronyms}.

\begin{table}[!ht]
\centering
\scriptsize
\caption{\MakeUppercase{List of Acronyms}}
\label{tab:acronyms}
\begin{tabular}{|c|l|}
\hline
A3C & Asynchronous Advantage Actor-Critic \\ \hline
A2C & Advantage Actor-Critic \\ \hline
ASA & Adversarial-Strategic Agent \\ \hline
AGE & Adversarially Guided Exploration \\ \hline
AI & Artificial Intelligence \\ \hline
ARPL & Adversarially Robust Policy Learning \\ \hline
ATN & Adversarial Transformation Networks \\ \hline
CARRL & Certified Adversarial Robustness for RL \\ \hline
C\&W & Carlini and Wagner \\ \hline
CDG & Common Dominant adversarial example Generation \\ \hline
c-MARL & Cooperative Multi-Agent Reinforcement Learning \\ \hline
DRL & Deep Reinforcement Learning \\ \hline
DDPG & Deep Deterministic Policy Gradient \\ \hline
DDQNs & Double Deep Q-Networks \\ \hline
d-JSMA & Dynamic budget JSMA \\ \hline
DL & Deep Learning \\ \hline
DQN & Deep Q-Networks \\ \hline
FGSM & Fast Gradient Sign Method \\ \hline
FRARL & Falsification-based RARL \\ \hline
GB & Gradient Based \\ \hline
GPS & Graded Policy Search \\ \hline
I2A & Imagination Augmented Agents \\ \hline
IRL & Inverse Reinforcement Learning \\ \hline
it-FGSM & Iterative target-based FGSM Method \\ \hline
JSMA & Jacobian-based Saliency Map Attack \\ \hline
KL & Kullback-Leibler \\ \hline
LAS & Look-ahead Action Space \\ \hline
MAD & Maximal Action Difference \\ \hline
MAS & Myopic Action Space \\ \hline
MBMF-RL & Model-Based priors for Model-free Reinforcement Learning \\ \hline
MDP & Markov Decision Process \\ \hline
ME-TRPO & Model Ensemble Trust Region Policy Optimization \\ \hline
ML & Machine Learning \\ \hline
MLAH & Meta-learned Advantage Hierarchy \\ \hline
MPC & Model Predictive Control \\ \hline
MuJoCo & Multi-Joint dynamics with Contact \\ \hline
MVE & Model-based Value Expansion \\ \hline
NAF & Normalized Advantage Function \\ \hline
NR-MDP & Noisy action Robust MDP \\ \hline
PCA & Principal Component Analysis \\ \hline
PEPG & Parameter Exploring Policy Gradients \\ \hline
POMDP & Partially Observable Markov Decision Process \\ \hline
PPO & Proximal Policy Optimization \\ \hline
PR-MDP & Probabilistic MDP \\ \hline
RARARL & Risk-Averse RARL \\ \hline
RARL & Robust Adversarial RL \\ \hline
RBFQ & Radial Basis Function based Q-learning \\ \hline
RNN & Recurrent Neural Network \\ \hline
SARSA & State-Action-Reward-State-Action Algorithm \\ \hline
SA-MDP & State-Adversarial MDP \\ \hline
SDN & Software-Defined Networking \\ \hline
SFD & Sampling-based Finite Difference \\ \hline
SGD & Stochastic Gradient Descent \\ \hline
SPG & Stochastic Policy Gradient \\ \hline
STEVE & STochastic Ensemble Value Expansion \\ \hline
TMDPs & Threatened Markov Decision Processes \\ \hline
TRPO & Trust Region Policy Optimization \\ \hline
WMA & Weighted Majority Algorithm \\ \hline
\end{tabular}
\end{table}

\section{Background}
\label{sec: background}

In this section, we discuss the fundamentals of the DRL process. Then, we provide a summary of the shortcomings of the ML and DRL techniques.

\subsection{Fundamentals of DRL}

The important concepts used in DRL are described as follows: 

\begin{itemize}

\item \textbf{Markov Decision Process:} A generic RL problem is described as an MDP in terms of the state, action, reward, and dynamics of the system. In an MDP, at each time step the agent observes the current state $s_t$ and performs an action $a_t$ based on its current policy $\pi^{*}$. After the action is executed, the agent observes its reward $r_t$ and next state $s_{t+1}$. The objective of an MDP is to find the best actions which maximize its long-term expected reward. Figure \ref{RL_basic} illustrates a typical MDP setup with an agent interacting with its surrounding environment.

\begin{figure}[!ht]
\centering
\includegraphics[width=0.35\textwidth]{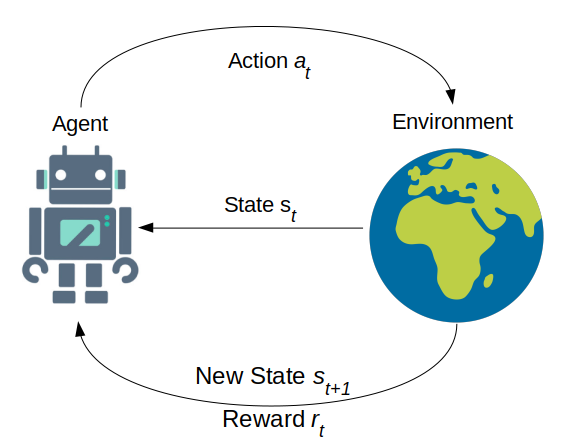}
\caption{Basic process of MDP in RL.}
\label{RL_basic}
\end{figure}

\item \textbf{Environment} is a simulator or a real-world scenario in which the agent interacts and learns. At each time step, the agent (governed by the policy) interacts with the environment and in return receives a reward. Environment is divided into two categories namely \textit{partially observable}, and \textit{fully observable}. In the case of a partially observable environment, the agent is only able to partially observe the environment. For these partially observable environments, POMDPs (Partially Observable MDPs) are used. In a fully observable environment, the agent can observe all the states and we use MDPs for this. MDPs are a special case of POMDPs where the observation function is identity.

\item \textbf{Action} is a stimulus used by the agent for interactions with the environment. The actions can be discrete or continuous based on the environment and DRL problem formulation. 

\item \textbf{Reward} is an incentive, expressed by a numerical value, that the agent receives after making an action. The goal of an agent is to maximize the accumulated reward. To reduce the impact of the reward $r$ which the agent might get in a later state due to taking a specific action $a$ in the current state $s_t$, the notion of discounted rewards was introduced. It is usually denoted by $\gamma$ and can take any value ranging from 0 to 1. Mathematically, the discounted reward $R_t$ given as:
\begin{equation} R_t = \sum_{t'=t}^{T} \gamma^{t'-t}r(s_{t'}), \label{value_r} \end{equation}

where $t$ denotes the timestep, $T$ is the final timestep, $r$ denotes the reward for the timestep, and $s_t$ denotes the current state.

\item \textbf{Value function} specifies the value of a state. Value is defined as the maximum expected discounted reward of a certain state. Mathematically, it is determined as:

\begin{equation} V_{\pi}(s) = \mathbf{E}_\pi[R_t | s=s_t], \end{equation}

where $\pi$ is the policy, $R_t$ is the discounted reward, and $s_t$ is the current state.

\begin{figure*}[!t]
\centering
\includegraphics[width=1\textwidth]{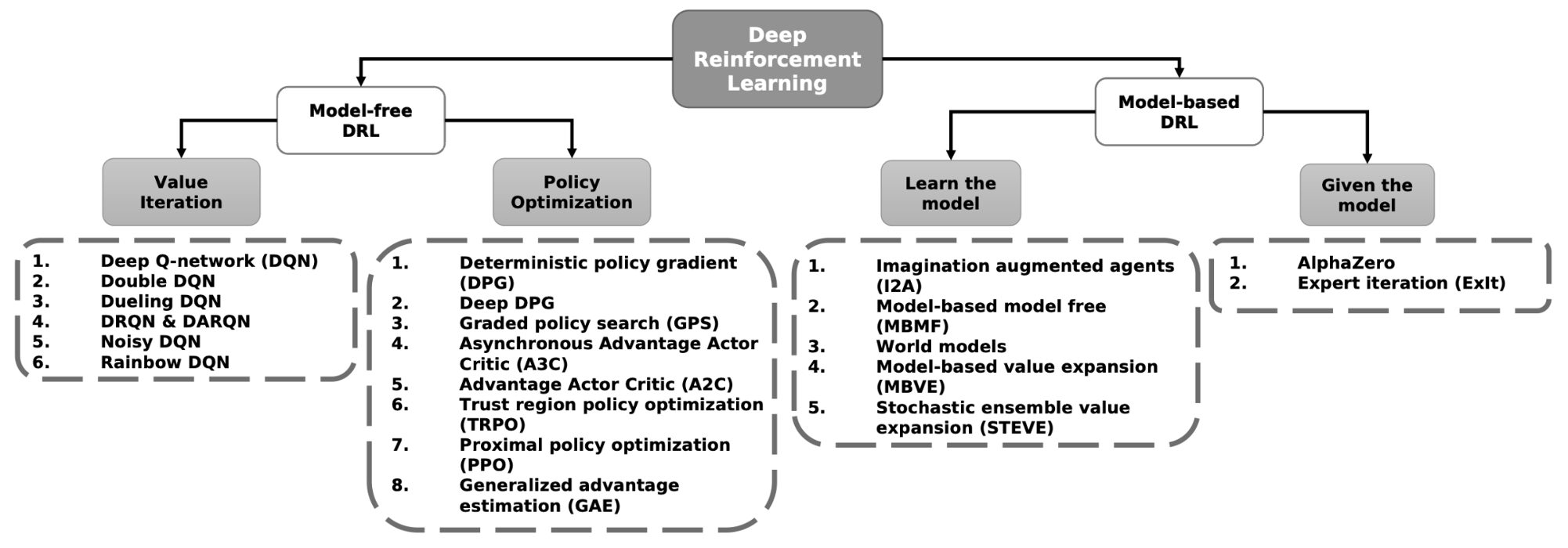}
\caption{A non-exhaustive taxonomy of major DRL schemes as proposed by \cite{OpenAISpinUp1}.}
\label{RL_DRL_taxonomy}
\end{figure*}

\item \textbf{Q-function} specifies the Q-value of a state. Q-value is defined as the maximum expected discounted reward an agent may get by taking a specific action at a specific state. Mathematically, it can be calculated as follows:

\begin{equation} Q_{\pi}(s) = \mathbf{E}_\pi[R_t | s=s_t,a=a_t], \end{equation}

where $\pi$ is the policy, $R_t$ is the discounted reward, $s_t$ is the current state, and $a_t$ is the current action.

\item \textbf{Advantage Function} is the difference of the Q-value of a specific action at a state $Q(s,a)$ from the value of that state $V(s)$, i.e., $A(s,a) = Q(s,a) - V(s)$.

\item \textbf{Policy} defines how the agent will behave in the environment at a particular time. In other words, it is a mapping from the perceived states of the environment to the actions taken in those conditions. A policy is said to be optimal if it achieves the maximum possible reward at each state. Policies are further divided into two types: namely, \textit{deterministic policy} and \textit{stochastic policy}. When actions taken by the agent are deterministic, the policy is termed as deterministic. On the other hand, when the actions are sampled from the conditional probability distribution of actions given the states, the policy is called to be stochastic.

\item \textbf{On-policy algorithms} enable an agent to learn and update its policy in an online manner through real-time interaction with the environment. Samples generated from the current policy are used to train the algorithm to estimate the policy in advance.

\item \textbf{Off-policy algorithms} use an online policy and a target policy. The target policy is used to estimate the action values while the online policy is being learned. Hence, the agent can estimate the target policy without its complete knowledge.

\item \textbf{Model} mimics the behavior of the environment, hence allowing inferences to be made about the behavior of the environment. Based on the availability of the system models, the DRL schemes are divided into further two categories namely \textit{model-based} and \textit{model-free RL}.

\item \textbf{Exploration and exploitation}: \textit{Exploration} is the process when the agent tries to explore the surrounding environment by taking different actions available at a given state. \textit{Exploitation} occurs after exploration. The agent exploits the optimal actions to achieve the maximum cumulative reward. An $\epsilon$-greedy policy is used to balance exploration and exploitation. The agent chooses a random action with a certain probability otherwise it takes the action followed by the policy. The probability of the random action being taken keeps decreasing with each timestep. This change factor is usually denoted by $\lambda$. 

\end{itemize}

A taxonomy of major DRL algorithms as proposed by \cite{OpenAISpinUp1} has been provided in Fig. \ref{RL_DRL_taxonomy}. We refer the interested readers to \cite{arulkumaran2017deep}, \cite{li2017deep} for further details on variations of DRL schemes.

\subsection{Security of ML}

Although the utilization of ML techniques has revolutionized many areas including vision, language, speech, and control; it has also introduced new security challenges that are very threatening in designing and developing new dynamic intelligent systems. Security attacks in ML can be divided into two categories namely training phase attacks and inference phase attacks. For the training phase attacks, the adversary tries to force the learning process to learn faulty model/policy by introducing small imperceptible perturbations to the input data. Inference phase attacks are performed by the adversary at the inference/test time of the ML pipeline to fool the model/policy in providing malfunctioned results/actions.

\begin{figure*}[!t]
\centering
\includegraphics[width=0.7\textwidth]{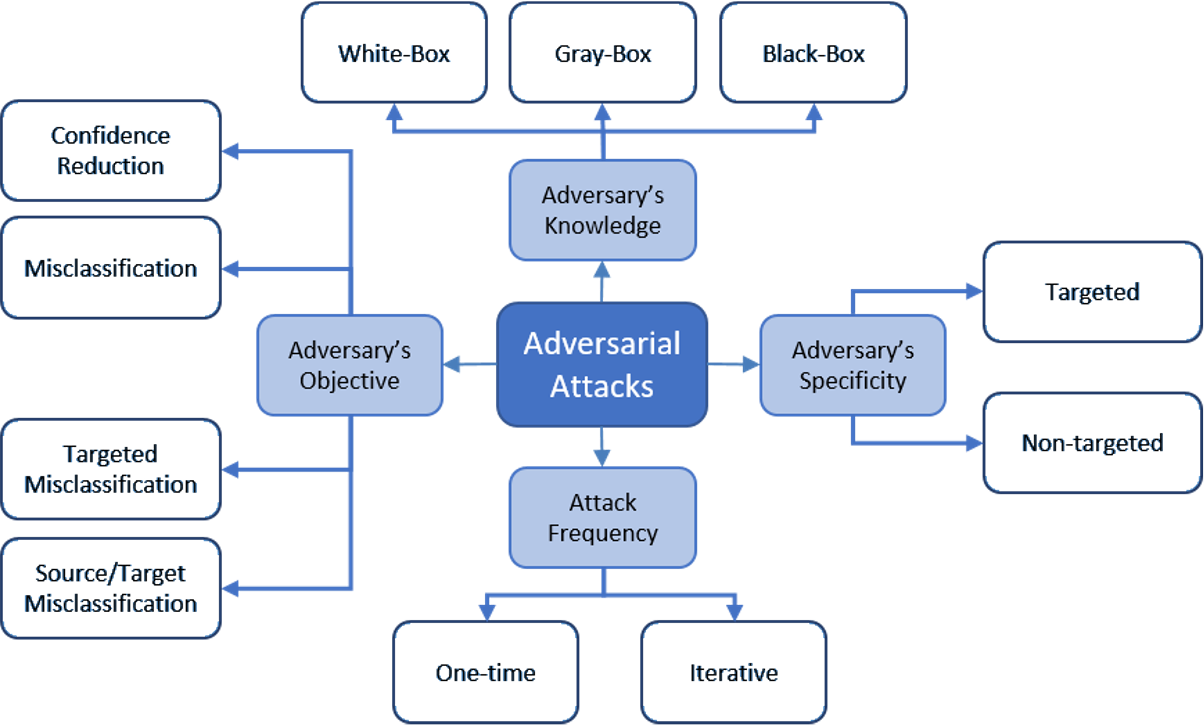}
\caption{Taxonomy of adversarial ML attacks classified according to the adversary’s objective, knowledge, specificity, and attack frequency.}
\label{attacks_ML}
\end{figure*}

The malicious input generated by adding adversarial perturbations into the original input is known as an \textit{adversarial example}. Adversarial examples are classified into four major categories based on the objective, knowledge, frequency, and specificity. Formally, an adversarial example $x^*$ is created by adding a small imperceptible carefully crafted perturbation $\delta$ to the correctly classified example $x$. The perturbation $\delta$ is calculated by approximating the optimization problem iteratively until the crafted adversarial example gets classified by ML classifier $f(.)$ in targeted class $t$ as follows:

\begin{equation}
 x^* = x + \arg \underset{\delta{_x}}{\text{min}} {\|\delta\|: f(x + \delta) = t},
 \label{equ1}
\end{equation}

where $t$ is the targeted class. Figure \ref{attacks_ML} shows a basic taxonomy of attacks on ML.

\subsubsection{\textbf{Attacks based on adversary's knowledge}}
Depending on the adversary's knowledge about the targeted ML model, adversarial attacks are divided into further three categories: namely white-box attacks, gray-box attacks, and black-box attacks. In the case of \textit{white-box attacks}, the adversary has perfect knowledge of the target ML/DL algorithm, i.e., the adversary knows the training and testing data, parameters of the model, etc. These attacks are used for the worst-case security malfunction analysis of an ML/DL system. In the case of \textit{gray-box attacks}, the adversary is supposed to have limited knowledge (knowledge about feature representation and optimization algorithms only) about the targeted ML/DL model. The adversary designs a surrogate model on the limited knowledge available and uses \textit{transferability} property \cite{tramer2017space} of the adversarial examples (where an adversarial example evading a classifier will evade other similar classifiers even if they are trained on another dataset) to evade the ML/DL based system. The attacker may also have limited test access to the model, i.e., it may be able to ask the model the output on some inputs. In the case of \textit{black-box attacks}, the adversary does not know the model or any of its attributes. The adversary can only query the systems for labels or confidence scores and develop an adversarial perturbation based on the feedback provided by the deployed ML/DL model.

\subsubsection{\textbf{Attacks based on adversary's goals}}
Based on the adversary's objective, adversarial attacks are divided into four types:
\begin{enumerate}
 \item \textit{Confidence reduction attacks} in which adversarial attacks are launched to compromise the confidence levels of the predictions of the deployed ML/DL based system;
 \item \textit{Misclassification attacks} in which adversarial attacks are launched for disturbing the classification boundary of any class to cause misclassification;
 \item \textit{Targeted misclassification attacks} in which adversarial attacks are launched to misclassify only a targeted class; 
 \item \textit{Source/target misclassification attacks} in which adversarial attacks are launched to force misclassification of a specific source class into a specifically targeted class.
\end{enumerate}

\subsubsection{\textbf{Attacks based on adversary's specificity}}
Based on specificity, adversarial examples can be classified into two types, i.e., \textit{targeted} and \textit{non-targeted}. These concepts are similar to the ones as in the case of the adversary's objective. In the case of targeted attacks, the attackers target specific classes in the output, while in the case of non-targeted attacks, the goal is to misclassify the maximum number of samples.

Although adversarial examples are transferable from one ML model to another but in many cases, the performance of the transferred examples is not enough. To further improve the performance of black-box attacks while reducing the number of queries needed for the attack, query-efficient black-box attacks are required. Different query-efficient black-box attack methods are available in the literature. Cheng et al.~\cite{cheng2018query} use randomized gradient-free methods for the creation of adversarial examples and show their algorithm to require 3-4 times fewer queries to achieve the same performance as the state-of-the-art attacks. Chen et al.~\cite{chen2017zoo} uses the zeroth-order optimization technique for adversarial perturbation generation and shows their black-box attack to demonstrate the same performance as state-of-the-art white-box attacks. The queries required by their technique \cite{chen2017zoo} are less than those required by that of \cite{cheng2018query}. The authors in \cite{tu2019autozoom} propose a more query-efficient attack. They propose autoencoder-based zeroth-order optimization for adversarial image generation in black-box attacks. They show a reduction of more than 93\% in the mean query count while maintaining the same performance as the state-of-the-art attacks. More details on adversarial ML can be found in \cite{zhang2019adversarial}, \cite{qayyum2019securing}.

\subsection{Security of DRL}

The increasing use of DRL in practical applications has led to an investigation of the security risks it faces. However, the security challenges faced by DRL are different from those experienced by other ML algorithms. The major difference is that a DRL process is trained to solve sequential decision-making problems in contrast to most other ML schemes that are trained to solve single-step prediction problems. The independence of the current actions from the previous ones increases the degrees of freedom of adversarial attacks raising new challenges that must be addressed. This makes the adversarial attacks more challenging to be recognized, as we cannot discriminate between the action intentionally taken by the agent and the action the adversary forces/lures the agent to take. Also, the training is done on a dataset from a fixed distribution in the case of ML, in contrast to the DRL where the agent begins with a deterministic or stochastic policy and starts exploring for best actions.

\begin{figure*}[!ht]
\centering
\includegraphics[width=0.8\textwidth]{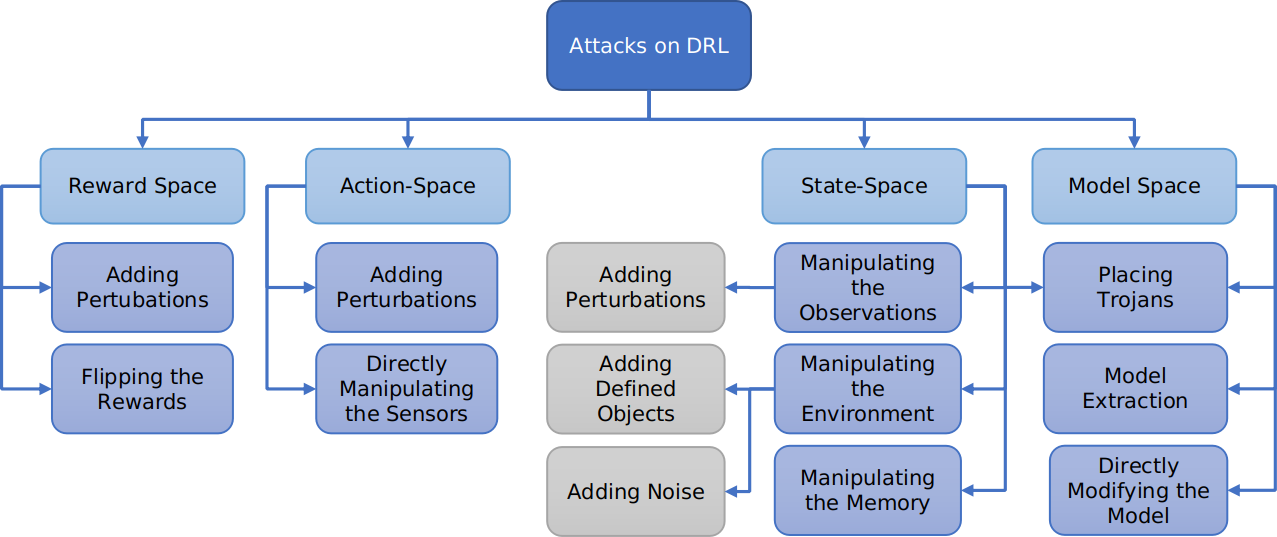}
\caption{Taxonomy of adversarial attacks on DRL classified according to the major parts of DRL.}
\label{DRL_attacks}
\end{figure*}

Usually, RL problems are formulated as an MDP consisting of four parts ($S$,$A$,$R$,$P$), where $S$ is the state space, $A$ is the action space, $R$ is the reward function and $P$ is the transition matrix between states. Hence, an adversary has more choices to attack. If the adversary targets the state space, imperceptible perturbations can be added to the environment directly by perturbing the sensors \cite{clark2018malicious}. Similarly, an adversary can target any of the four major components of MDP. Adversarial attacks on DRL are classified into \textit{inference-time} and \textit{training-time} attacks \cite{vorobeychik2018adversarial}. An adversary may compromise one or more than one dimension of confidentiality, integrity, and availability \cite{behzadan2018faults}.

Based on the goal of the adversary, the adversarial attacks on DRL can be classified into \textit{active} or \textit{passive} \cite{behzadan2018faults}. For active attacks, the adversary desires to change the behavior of the agent, while for passive attacks, the adversary desires to infer details about the model, reward function, or other parts of DRL. An adversary can use these details to either create a copy of the model or use them to perform an attack on the model. The adversary may be limited by the part of the environment, where an adversary is only capable of making changes to a certain area of the environment. Adding a lot of perturbation in a single time instance may make the attack perceptible which is not preferred by the adversary. Distinguishing the adversarial samples and behavior from the normal ones in the case of DRL is not as easy as in supervised learning because of the increased possible attack dimensions.

\section{Adversarial Attacks on DRL}
\label{sec:attacks}

In this section, we discuss the adversarial attacks on DRL. We divide the attacks on DRL into four categories based on the functional components of the DRL process. A major portion of the attacks involve the addition of adversarial perturbations to the state space and a small portion of the proposed attacks involve perturbing the reward and action space. Figure \ref{DRL_attacks} shows a basic taxonomy of the adversarial attacks on DRL algorithms.\par

\begin{figure}[!t]
\centering
\includegraphics[width=0.48\textwidth]{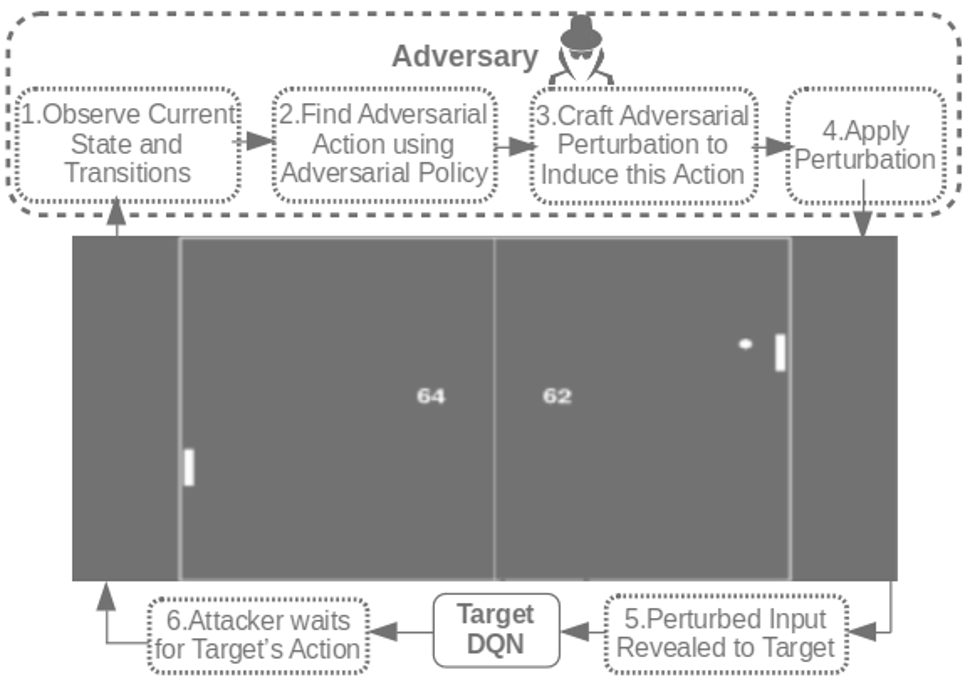}
\caption{The process of policy induction attack \cite{behzadan2017vulnerability} performed on the game of Pong.}
\label{fig:behzadan_pi}
\end{figure}

\subsection{Attacks Perturbing the State space}

\begin{figure*}[!t]
\centering
\includegraphics[width=0.80\textwidth]{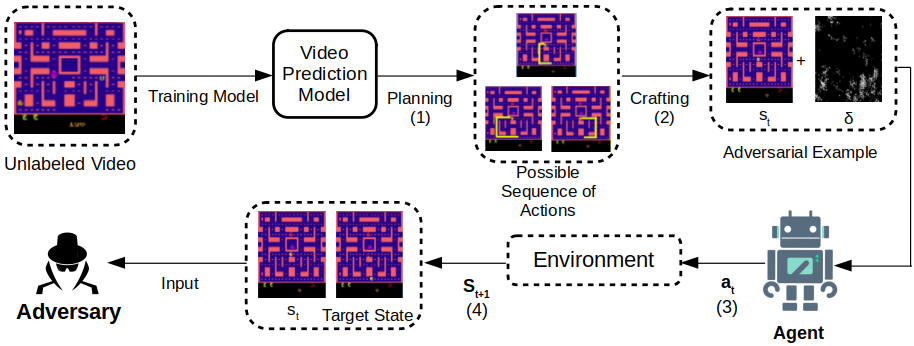}
\caption{An illustration of enchanting attack on Pacman is depicted which is highlighting all four components: (1) action sequence planning, (2) crafting an adversarial example with a target-action, (3) the agent takes an action, and (4) environment generates the next state $s_{t+1}$.}
\label{fig:lin}
\end{figure*}

We divide this subsection based on the access of the adversary.

\subsubsection{Manipulating the Observations}
Since DNNs are vulnerable to adversarial attacks in supervised learning, we would expect DNNs trained via deep RL to also be vulnerable. Indeed, Behzadan and Munir \cite{behzadan2017vulnerability} show this vulnerability and verify the transferability of adversarial examples across different DQN models. They consider a man-in-the-middle adversary between the environment and the DRL agent, where the adversary perturbs the states from the environment and forwards these perturbed states to the DRL agent to take the desired action. To ensure the imperceptibility of the perturbation, the amplitude of the adversarial examples crafting algorithms (Fast Gradient Sign Method (FGSM) and Jacobian-based Saliency Map Attack (JSMA) \cite{papernot2016limitations}) is controlled. The attack procedure is divided into two phases, initialization and exploitation. The \textit{initialization} phase includes the training of a DQN on adversarial reward function to generate an adversarial policy. Then, a replica of the target's DQN is created and initialized from random parameters. The \textit{exploitation} phase includes generating adversarial inputs such that the target DQN can be made to follow actions governed by the adversarial policy. Furthermore, they propose an attack method to manipulate the policy of the DQN by exploiting the transferability of adversarial samples. They use a black-box setting and show a success rate of 70\% when adversarial examples are transferred from one model to another. The cycle of the proposed policy induction attack is shown in Figure \ref{fig:behzadan_pi}. Huang et al.~\cite{huang2017adversarial} use the attack proposed in \cite{behzadan2017vulnerability} and show a significant drop in the performance of DQN, \textit{trust region policy optimization} (TRPO), and \textit{asynchronous advantage actor-critic} (A3C) methods in both white and black-box settings. They show the DQN to be more susceptible to adversarial attacks than the TRPO and A3C. 

Lin et al.~\cite{lin2017tactics} argue that the uniform attack schemes may not be practically feasible and are easy to detect. They consider a different approach and propose two adversarial attack techniques on DRL schemes, namely, \textit{strategically-timed attack} and \textit{enchanting attacks}. For the former, they propose to minimize the reward of the DRL schemes by using adversarial examples on a subset of time steps in an episode of the DRL operation. For the latter, they propose a novel method of luring the DRL agent to a predefined targeted state by using a generative model and a sophisticated planning algorithm. The generative model is used to predict the next state in the space and the planning algorithm is used to generate actions required for luring the agent to the targeted state. Performance of the strategically-timed attack and the enchanting attack is reported on DQN and A3C agents, playing Atari games where 70\% of the success rate of the adversarial attacks is reported. They use the Carlini and Wagner (C\&W) attack \cite{carlini2017towards} for generating adversarial inputs. It is also shown that perturbing only 25\% of the inputs using the proposed method produces the same results as the previously proposed attacks based on FGSM. The workflow of the enchanting attack is shown in Figure \ref{fig:lin}.

Tretschk et al.~\cite{tretschk2018sequential} propose a similar approach to the enchanting attacks proposed by Lin et al.~\cite{lin2017tactics} where they use the adversarial transformer network (ATN) \cite{baluja2018learning} to impose adversarial reward on the policy network of DRL. The ATN makes the agent maximize the adversarial reward through a sequence of adversarial inputs. Complete information regarding the agent and the target environment is required, hence making the attack white box. It is shown that given a large enough threshold for perturbation, the agent can be made to follow the adversarial policy at the test time. 

Pattanaik et al.~\cite{pattanaik2018robust} prove that FGSM-based attacks on DRL \cite{huang2017adversarial} do not use an optimal cost function for crafting the adversarial inputs and propose a loss function that is guaranteed to maximize the probability of taking the worst possible action. They propose three types of gradient-based adversarial attacks on DQN and \textit{deep deterministic policy gradients} (DDPG) techniques for reducing the expected reward by adding perturbations to the observations. The \textit{first attack} is based on a naive approach of adding random noise to the DRL states to mislead the DRL agent in selecting a sub-optimal action that decays the performance of the DRL scheme. The \textit{second attack} is a gradient-based (GB) attack, where a new cost function is introduced for creating adversarial actions, that outperforms the FGSM in determining the worst possible discrete action to limit the performance of DRL schemes. The \textit{third attack} is an improved version of the second attack. Instead of using a simple gradient-based approach for generating adversarial perturbation, the authors use stochastic gradient descent (SGD) for adversarial action generation which ultimately misleads the DRL agent to end up in a pre-defined adversarial state. 

Kos and Song \cite{kos2017delving} discuss that the previous attacks require perturbing several states to be successful which may not be practically feasible. They propose to use a value function to guide the adversarial perturbation injection hence reducing the number of adversarial perturbations needed for introducing a malfunction in DRL policies. They propose three types of attack situations: (1) the addition of noise at a fixed frequency; (2) the addition of specially designed perturbed inputs after $N$ samples; (3) the recalculation of the perturbation after $N$ samples and adding the previously calculated perturbation to the intermediate steps. The results show their last approach performs as well as the one in which all states are perturbed. Furthermore, they use the generated samples for retraining the model and show that resilience can be improved against both FGSM and random adversarial perturbations.

A similar issue has been raised by Sun et al.~\cite{sun2020stealthy}. Furthermore, they discuss that the previously proposed attacks are not general-purpose and have limitations, e.g., \cite{lin2017tactics} cannot be used for continuous action spaces. They propose two sample efficient general-purpose attacks that can be used to attack any DRL algorithm while considering long-term damage impacts, namely: \textit{critical point attack} and \textit{antagonist attack}. The first one involves the building of a model by the adversary to predict future environmental states and the agent’s actions. The damage of each possible attack strategy is then assessed and the optimal one is chosen. The antagonist attack involves automatic learning of a domain-agnostic model by the adversary to discover the critical moments of attacking the agent in an episode. To be successful, the critical point technique only requires 1 (TORCS) or 2 (Atari pong and breakout) steps, and the antagonist technique needs fewer than 5 steps (4 MuJoCo tasks).

Hussenot et al.~\cite{hussenot2019targeted} discuss that the previously proposed approaches are either practically infeasible or computationally extensive. They propose two types of adversarial attacks to take full control of the DRL agents' policy. The first one called \textit{per-observation attack} includes the generation of a new adversarial perturbation for every observation of the agent and adding that perturbation to the environment. The second one called \textit{universal mask attack} includes the addition of one universal perturbation, created at the start of the attack, to all the observations. These attacks are discussed in both targeted and non-targeted situations. It is also reported that the proposed attacks are more successful if the FGSM is used for generating the perturbations in untargeted attack situations, whereas in the case of targeted attacks the FGSM is not able to generate imperceptible adversarial samples.

Chan et al.~\cite{chan2020adversarial} take a different approach from other articles and propose static reward impact maps which can be used to quantify the influence on the reward of each feature in the state space. By the use of such a map, the adversary can choose to perturb only those features which have a large impact on the cumulative reward. The time complexity of the generation of these static maps is posed as a limitation.

Cooperative Multi-Agent Reinforcement Learning (c-MARL) algorithms are gaining attention in a wide range of applications such as cellular base station control \cite{de2018cooperative}, traffic light control \cite{wiering2000multi}, and autonomous driving \cite{shalev2016safe}. The target of the agent in c-MARL is to learn to take action cooperatively as a team to maximize a total team reward. Lin et al.~\cite{lin2020robustness} show the vulnerability of c-MARL agents to adversarial attacks by proposing a mechanism of adding perturbations to the state space. Difficulty to estimate team rewards; difficulty to measure the effect of misprediction of an agent on the team reward; non-differentiability of models; and low-dimensionality of the feature space, make attacking such environments challenging. They hypothesize that the cooperative aspects of c-MARL agents make these agents more vulnerable to adversarial attacks as compared to single-agent RL, as the failure of a single agent may cause the failure of the whole team. They extend the FGSM attack and propose two new approaches to decrease the team reward more effectively: namely, \textit{Iterative target-based FGSM method (it-FGSM)} and \textit{Dynamic budget JSMA attack (d-JSMA)}. These attacks involve training an adversarial policy network to search for a sub-optimal action from which the adversarial examples are then introduced in the observations of one of the agents to force it to take the targeted action. They test their attack on the StartCraft II multi-agent benchmark. They show that their attack can decrease the reward from 20 to 9.4 by attacking only a single agent out of the multiple possible agents when the perturbations are added with an average of 8.1 L1 norm. As a reaction to this drop in reward, the winning rate of the multi-agent drops from 98.9\% to 0\%. Furthermore, they discuss the applicability of the proposed attack in real environments as an adversary can gain access to a single agent and use it to attack the whole system.

Despite the success of DRL, there is little research that studies the impact of adversarial attacks in DRL algorithms that do not use images as inputs. Wang et al.~\cite{wang2020adversarial} propose techniques that can degrade the performance of a well-trained DRL-based energy management system of an electric vehicle causing them to either use much fuel or lead it into running out of battery before the end of the trip. They show their adversarial inputs to be imperceptible. For white-box settings, they use FGSM as the adversary while for black-box settings, attack transferability, and finite-difference method \cite{xiao2019characterizing} are used. They test their attacks on a DQN trained for energy management of an electric vehicle and show the degradation of performance.

\subsubsection{Manipulating the Environment}

Adversarial attacks on the state space can also be carried out by adding perturbations in the environment of the agent. In turn, this causes the agent to consider the environment as the adversary desires. Chen et al.~\cite{chen2018gradient} propose \textit{common dominant adversarial examples generation method} (CDG) for crafting adversarial examples with high confidence for the environment of DRL. The core idea of their attack is the addition of confusing obstacles to the original clean map for the case of pathfinding to confuse the robot by messing with its local information. For a perturbation to be successful, it should either stop the agent from reaching the destination or otherwise delay the agent. The proposed attack is tested on A3C and is shown to be successful at least 99.91\% of the time.

Bai et al.~\cite{bai2018adversarial} take a different approach than \cite{chen2018gradient} and propose a method of finding adversarial examples for DQNs trained for automatic pathfinding. The proposed attack analyzes a trained DQN for the task and identifies the weaknesses present in the Q-value curves. Specially designed perturbations are added to these weaknesses in the environment to effectively refrain the agent from learning an optimal solution to the maze.

Xiao et al.~\cite{xiao2019characterizing} introduce online sequential attacks on the environment of the DRL agent by exploiting the temporal consistency of the states. This attack performs faster than the FGSM algorithm as no back-propagation is needed and is based on model querying. The authors provide two methods for model querying, namely \textit{adaptive dimension sampling-based finite difference method} (SFD), and \textit{optimal frame selection method}. In addition to these sequential attacks, they also propose other attacks on the observations, action selection, and environment dynamics. 

Gleave et al.~\cite{gleave2019adversarial} propose to introduce an adversarial agent in the same environment as the legitimate agent. The adversary is not able to manipulate the observations of the legitimate agent but can create natural observations that can act as adversarial inputs and make the agent follow the target policy. This leads to a zero-sum game between the adversarial agent and the legitimate agent. Both the adversarial and the victim agent are based on PPO. After showing the existence of such adversarial policies, they suggest that the learning of the deployed model must be frozen to save them from undesired behaviours enforced by adversaries. Such adversarial agents can also be used in making the models better by constantly attacking and retraining.

Although the DRL techniques work better in navigation tasks, they are more vulnerable to adversarial attacks than the classic methods. Yang et al.~\cite{yang2020enhanced} introduce timing-based adaptive adversarial perturbations for learning-based systems in real-world scenarios. They introduce two attacks namely: \textit{weighted majority algorithm} (WMA) (white-box setting) and \textit{adversarial-strategic agent} (ASA) via a population-based training method based on \textit{parameter exploring policy gradients} (PEPG) (black-box setting). The first one is based on online learning while the other is based on evolutionary learning. Through experiments, they show that out of the two proposed attacks, WMA shows a better performance.

\subsubsection{Manipulating the Training Data}

The adversary can also choose to perturb the training data to indirectly having the agent follow a targeted policy.
An adversary may create and hide some deficiencies in the policy to use them later for his benefit. Kiourti et al.~\cite{kiourti2019trojdrl} show this vulnerability of DRL models to Trojan attack with adversary having access to the training phase of the model. It is reported that by only modifying 0.025\% of the training data, an adversary can induce such hidden behaviors in the policy that the models perform perfectly well until the Trojan is triggered. The proposed attack is shown to be resistant against current defense techniques for Trojans. 

A similar approach has been proposed by Behzadan and Hsu \cite{behzadan2019sequential} to secure DRL models from model extraction attacks but can be used for adversarial purposes. This involves the integration of a unique response to a specific sequence of states while keeping its impact on performance minimum, hence saving from the unauthorized replication of policies. It is also shown that the unwatermarked policies are not able to follow the identified trajectory which is specified during the training. As already discussed, this can be used by adversaries to hide specific patterns in the policy and use them to their benefit later.

\subsubsection{Manipulating the Sensors}

The research on real-time attacks on robotic systems in a dynamic environment has not been much explored. Clark et al.~\cite{clark2018malicious} evaluate a white-box adversarial attack on the DRL policy of an autonomous robot in a dynamic environment. The goal of the DRL robot is to reach the destination by routing through the environment, while the goal of the adversary is to mislead the agent into the wrong routes. The adversary misleads the agent into following false routes by tampering with sensory data. They also observe that once the adversarial input is removed, the robot automatically reverts to taking the correct route. Hence, an attacker can modify the behavior of the model temporarily and leave behind zero or very little evidence. Their attack requires access to the trained policy but not the hyper-parameters used during training.

A similar observation has been shown by Usama et al.~\cite{usama2020examining}. They argue that a lot of research is being done for creating AI/ML solutions to problems in future networks, such as IoT and 6G. They show these ML systems to be vulnerable by highlighting the adversarial dimension of these systems. They prove their point by attacking a DRL-based channel auto-encoder framework and showing its drop in performance. Noise is added to the feedback channel for a certain time interval. Furthermore, they show that when this noise is removed, the DRL system automatically can regain its original performance, leaving no footsteps by the adversary.

\subsection{Attacks Perturbing the Reward function}

Han et al.~\cite{han2018reinforcement} discuss the reaction of the DRL agent in software-defined networking (SDN) to different adversarial attacks. The adversary adopts white-box and black-box settings for both inference and poisoning attacks in an online setting. They propose two types of attacks: \textit{flipping reward signals} and \textit{manipulating states}. For flipping reward signals, the adversary can manipulate the binary reward signal of the model by flipping it a certain number of times. For state manipulation, the attacker makes two changes in the first few steps of the training, i.e., an adversary can change the binary reward of two states from 0 to 1 and 1 to 0, respectively. Hence, the adversary can change the label of one compromised node to be uncompromised and vice versa.

A similar falsification approach has been followed by Huang and Zhu \cite{huang2019deceptive} leading the agent into taking targeted decisions. They characterize a robust region for policy in which the adversary can never achieve the desired policy while keeping the cost in this region. They use four terms to specify different types of attackers: namely (1) \textit{omniscient attacker} who has all the information before a certain time $t$; (2) \textit{peer attacker} who does not know about the transition probabilities but has access to the knowledge the agent has before a time $t$; (3) \textit{ignorant attacker} who only knows the cost signals before a time $t$; and (4) \textit{blind attacker} that has no information at time $t$. All these attackers may be limited by the budget of the attack and other constraints. It is shown that by the falsification of the cost at each state, all of these adversaries can mislead the agent into learning a policy desired by the adversary.

Rakhsha et al.~\cite{rakhsha2020policy} propose a training-time attack involving the poisoning of the learning environment to force the agent into executing a target policy chosen by the adversary. They consider RL agents that maximize average reward in undiscounted infinite-horizon settings and argue this to be a more suitable objective for many real-world applications that have cyclic tasks or tasks without absorbing states, e.g., a scheduling task and an obstacle-avoidance robot. They suppose that the adversary can manipulate the rewards and the transition dynamics in the learning environment at training-time in a stealthy manner. They test their attack in both offline and online settings. In the former, the agent is planning in a poisoned environment while in the latter, the agent is learning a policy using a regret-minimization framework with poisoned feedback. They show that the adversary can easily succeed in teaching an adversarial policy to the RL agent.

\subsection{Attacks Perturbing the Action space}

The adversary can have access to the actuators and might try to perturb the actions taken.
Yeow et al.~\cite{yeow2019spatiotemporally} propose two attacks on the action space of the DRL algorithms. The first one is an optimization problem for minimizing the cumulative reward of the DRL agent with decoupled constraints called \textit{myopic action space (MAS)} attack. The second one has the same objective as the first one but with temporally coupled constraints called \textit{look-ahead action space (LAS)} attack. The results show that LAS is more lethal in deteriorating the performance of the DRL algorithm as it can attack the dynamic information of the agent. This attack is also shown to perform well in the case of limited resources. Such attacks can be used to gain insights into the potential vulnerabilities of the DRL model. They also speculate that their proposed attacks on reward signals by perturbing the action space cannot be defended as the action space is independent of the policy. However, it can be detected by having a look at the decay in the reward.

\begin{table*}[!ht]
\caption{\MakeUppercase{Summary of the adversarial attacks on DRL pipeline highlighting the threat model and attack location in the DRL pipeline}}
\centering
\label{tab:attacks}
\resizebox{\textwidth}{!}{%
\begin{tabular}{|c|c|c|c|c|c|c|c|c|c|c|}
\hline
\multirow{2}{*}{\textbf{Paper}} & \multirow{2}{*}{\textbf{DRL Technique Targeted}} & \multirow{2}{*}{\textbf{Environment}} & \multicolumn{2}{c|}{\textbf{Test-time Attacks}} & \multicolumn{2}{c|}{\textbf{Train-time Attacks}} & \multicolumn{4}{c|}{\textbf{Attack Target}} \\ \cline{4-11} 
 & & & \textbf{White Box} & \textbf{Black Box} & \textbf{White Box} & \textbf{Black Box} & \textbf{State} & \textbf{Action} & \textbf{Reward} & \textbf{Model} \\ \hline
Behzadan and Munir~\cite{behzadan2017vulnerability} & DQN & pong & \cmark & \cmark & \xmark & \cmark & \cmark & \xmark & \xmark & \xmark \\ \hline
Huang et al.~\cite{huang2017adversarial} & DQN, TRPO \& A3C & \begin{tabular}[c]{@{}c@{}}chopper command, pong, \\ seaquest, space invaders\end{tabular} & \cmark & \cmark & \xmark & \cmark & \cmark & \xmark & \xmark & \xmark \\ \hline
Kos and Song~\cite{kos2017delving} & A3C & pong & \cmark & \xmark & \xmark & \xmark & \cmark & \xmark & \xmark & \xmark \\ \hline
Pattanaik et al.~\cite{pattanaik2018robust} & DQN \& DDPG & cartpole, mountain car & \cmark & \xmark & \xmark & \xmark & \cmark & \xmark & \xmark & \xmark \\ \hline
Lin et al.~\cite{lin2017tactics} & DQN \& A3C & \begin{tabular}[c]{@{}c@{}}pong, seaquest, mspacman, chopper\\ command, qbert\end{tabular} & \cmark & \xmark & \xmark & \xmark & \cmark & \xmark & \xmark & \xmark \\ \hline
Tretschk et al.~\cite{tretschk2018sequential} & DQN & Pong & \cmark & \xmark & \xmark & \xmark & \cmark & \xmark & \xmark & \xmark \\ \hline
Clark et al.~\cite{clark2018malicious} & DQN & Pathfinding & \cmark & \xmark & \xmark & \xmark & \cmark & \xmark & \xmark & \xmark \\ \hline
Chen et al.~\cite{chen2018gradient} & A3C & Pathfinding & \xmark & \xmark & \cmark & \xmark & \cmark & \xmark & \xmark & \xmark \\ \hline
Han et al.~\cite{han2018reinforcement} & DDQN \& A3C & Software-defined networking & \cmark & \cmark & \cmark & \xmark & \xmark & \xmark & \cmark & \xmark \\ \hline
Behzadan and Hsu~\cite{behzadan2019adversarial} & DQN, A2C \& PPO & Cartpole & \xmark & \cmark & \xmark & \xmark & \xmark & \xmark & \xmark & \cmark \\ \hline
Kiourti et al.~\cite{kiourti2019trojdrl} & A2C & \begin{tabular}[c]{@{}c@{}}Pong, Space Invaders, Qbert,\\ Breakout, Seaquest, Crazy\\ Climber\end{tabular} & \xmark & \xmark & \cmark & \cmark & \cmark & \xmark & \xmark & \xmark \\ \hline
Yeow et al.~\cite{yeow2019spatiotemporally} & PPO and DDQN & Lunar-Lander, BiPedal Walker & \xmark & \xmark & \cmark & \xmark & \xmark & \cmark & \xmark & \xmark \\ \hline
Hussenot et al.~\cite{hussenot2019targeted} & DQN \& Rainbow DQN \cite{hessel2018rainbow} & \begin{tabular}[c]{@{}c@{}}pong, space invaders, air raid,\\ and HERO\end{tabular} & \cmark & \xmark & \cmark & \xmark & \cmark & \xmark & \xmark & \xmark \\ \hline
Xiao et al.~\cite{xiao2019characterizing} & DQN \& DDPG & \begin{tabular}[c]{@{}c@{}}TORCS \cite{pan2017virtual}, \\ Atari (pong, enduro),\\ MuJoCo (half-cheetah, hopper)\end{tabular} & \cmark & \cmark & \xmark & \xmark & \cmark & \cmark & \xmark & \xmark \\ \hline
Huang and Zhu~\cite{huang2019deceptive} & Q-Learning & water reservoir system & \xmark & \xmark & \cmark & \xmark & \xmark & \xmark & \cmark & \xmark \\ \hline
Behzadan and Hsu~\cite{behzadan2019sequential} & DQN & cartpole & \xmark & \xmark & \cmark & \xmark & \cmark & \xmark & \xmark & \xmark \\ \hline
Bai et al.\cite{bai2018adversarial} & DQN & Pathfinding & \xmark & \xmark & \cmark & \xmark & \cmark & \xmark & \xmark & \xmark \\ \hline
Gleave et al.~\cite{gleave2019adversarial} & PPO & \begin{tabular}[c]{@{}c@{}}MuJoCo: kick and defend,\\ you shall not pass,\\ sumo humans, sumo ants\end{tabular} & \xmark & \cmark & \xmark & \xmark & \cmark & \xmark & \xmark & \xmark \\ \hline
Sun et al.~\cite{sun2020stealthy} & A3C, DDPG, \& PPO & \begin{tabular}[c]{@{}c@{}}TORCS, Atari (pong and breakout),\\ and MuJoCo (inverted pendulum,\\ half-cheetah, hopper, and walker2D)\end{tabular} & \cmark & \cmark & \xmark & \xmark & \cmark & \xmark & \xmark & \xmark \\ \hline
Lin et al.~\cite{lin2020robustness} & c-MARL & StartCraft II & \xmark & \cmark & \xmark & \xmark & \cmark & \xmark & \xmark & \xmark \\ \hline
Yang et al.~\cite{yang2020enhanced} & DQN \& A3C & \begin{tabular}[c]{@{}c@{}}reacher, banana collector, and\\ donkey car\end{tabular} & \xmark & \xmark & \cmark & \cmark & \cmark & \xmark & \xmark & \xmark \\ \hline
Raksha et al.~\cite{rakhsha2020policy} & RL & obstacle-avoidance, scheduling & \xmark & \xmark & \cmark & \xmark & \xmark & \xmark & \cmark & \xmark \\ \hline
Wang et al.~\cite{wang2020adversarial} & DQN & \begin{tabular}[c]{@{}c@{}}Energy management system of\\ an electric vehicle\end{tabular} & \cmark & \cmark & \xmark & \xmark & \cmark & \xmark & \xmark & \xmark \\ \hline
Chen et al.~\cite{chen2020stealing} & \begin{tabular}[c]{@{}c@{}}DQN, PPO, ACER \cite{wang2016sample}, \\ ACKTR \cite{wu2017scalable} \& A2C\end{tabular} & cartpole and pong & \xmark & \cmark & \xmark & \xmark & \xmark & \xmark & \xmark & \cmark \\ \hline
Huai et al.~\cite{huai2020malicious} & A3C, DQN & pong, breakout, space invaders & \cmark & \xmark & \cmark & \xmark & \cmark & \xmark & \xmark & \cmark \\ \hline
Chan et al.~\cite{chan2020adversarial} & DQN, PPO & \begin{tabular}[c]{@{}c@{}}freeway, pong, breakout,\\ fishing-derby, space-invaders,\\ battlezone\end{tabular} & \cmark & \cmark & \xmark & \xmark & \cmark & \xmark & \xmark & \xmark \\ \hline
Usama et al.~\cite{usama2020examining} & Basic DRL & SDN & \xmark & \cmark & \xmark & \xmark & \cmark & \xmark & \xmark & \xmark \\ \hline
Lee et al.~\cite{lee2020query} & PPO & point goal, car goal \cite{ray2019benchmarking} & \xmark & \cmark & \xmark & \cmark & \xmark & \cmark & \xmark & \xmark \\ \hline
\end{tabular}%
}
\end{table*}

Due to the difficulty of obtaining the complex models for cyber-physical systems for traditional control, they are being shifted to DRL. Lee et al.~\cite{lee2020query} argue that before transferring these systems to DRL, the security limitations of DRL must be understood. They propose a query-based attack for perturbing the action space of DRL in such systems. Furthermore, they show that by the use of adversarial training the attack success can be reduced to half.

\subsection{Attacks Perturbing the Model Space}

The adversary can have access to the model during or after training. Based on this access, the adversary might try to manipulate the model into learning the adversarial behavior or might also try to extract the learned model and use it later for attack purposes.
Behzadan and Hsu \cite{behzadan2019adversarial} propose an adversarial attack for targeting the confidentiality of the DRL policy. The proposed attack performs a model extraction attack by using imitation learning while querying the original model iteratively. They show that the adversarial examples generated for the model extracted are transferred successfully to the original model hence affecting its performance in a black-box setting. They use FGSM for generating adversarial examples for the imitated model. It is also shown that by providing the attack a sufficient number of observations, adversarial examples can be crafted with high efficiency. They use adversarial regret, i.e., the difference between maximum return achievable by the trained policy $\pi$ and return achieved from actions of adversarial policy, as a metric to measure the performance of their attacks. They show an increase in adversarial regret in case of an adversarial policy.

Chen et al.~\cite{chen2020stealing} argue that the techniques used for model extraction in supervised ML cannot be applied to RL due to high complexity and limited observable information and propose a technique for model extraction in DRL. At first, they use a recurrent neural network (RNN) classifier to reveal the training algorithm of the target black-box DRL model based on the predicted actions. Then, they use imitation learning to replicate the victim model from the extracted algorithm. A PPO is used for imitation learning. The extraction of models can be used by adversaries to generate successful adversarial examples making deployed models even more vulnerable to adversarial attacks.

Huai et al.~\cite{huai2020malicious} propose an optimization framework for deriving optimal adversarial attack strategy for model poisoning attacks. They propose two attacks: one in which adversarial perturbations are added to the observations of the agent and the other in which the attacker modifies the parameters of trained models in such a way that their performance is not affected. The first one is termed as \textit{universal adversarial attack against DRL interpretations} (UADRLI) while the latter is termed as a \textit{model poisoning attack against DRL interpretations} (MPDRLI). For UADRLI, they assume that the adversary has access to a certain area of the images (states) and cannot perturb pixels outside this certain area. The perturbations are only added at some time steps.

\subsection{Discussion}

\begin{figure*}[!ht]
\centering
\includegraphics[width=0.7\textwidth]{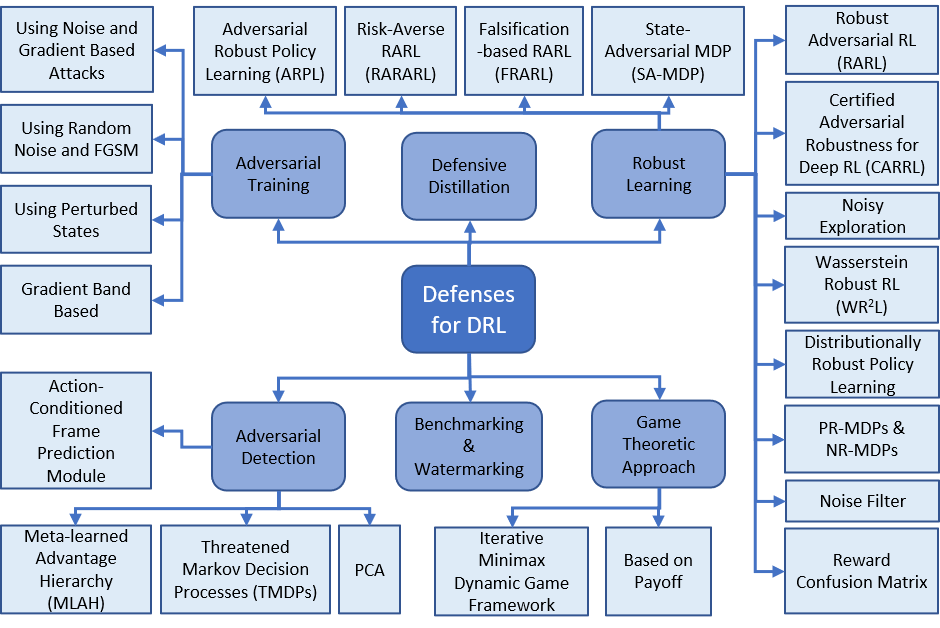}
\caption{A taxonomy of the major defense schemes used in DRL.}
\label{DRL_defenses}
\end{figure*}

In this section, we discuss the attacks on DRL by categorizing them based on the targeted part of the MDP. The adversary can target the state space, action space, reward function, or the model space based on the access available to the adversary. When targeting the state space, the adversary can add perturbations to the environment, training data, observations, and sensory data. In the case of perturbing the action space, the adversary can target the actuators. In the case of perturbing the reward function, the adversary can perturb the reward signal or might flip it. In the case of model-space attacks, the adversary can perturb the learned parameters of the model or might attempt to extract the learned model which might be proprietary, i.e., owned and copyrighted by some organization.

In real environments, the attacks that generate imperceptible and natural perturbations are more practical than the attacks that involve adding specially designed perturbations to states. In applications like autonomous driving, getting direct access to the sensors might not be possible for the adversary. The only option is to perturb the environment hence indirectly affecting the observations, actions, rewards, and policies. The real environments are often black-box where the adversary has no knowledge of the system being attacked and the number of queries is limited. The adversary has to improvise to attack the system where the target of the adversary can be to cause a drop in performance of the system or to evade the system. This puts forward a need for query-efficient attacks, similar to those proposed in \cite{tu2019autozoom} for supervised ML, to be proposed for DRL.

Table \ref{tab:attacks} shows a summary of the adversarial attacks on DRL.

\section{Defenses against adversarial attacks on DRL}
\label{sec:adv_ML_def}

In this section, we provide a detailed review of the countermeasures proposed to deal with adversarial attacks on DRL. Figure \ref{DRL_defenses} shows a basic taxonomy of the defenses that can be used for securing DRL algorithms.\par

\subsection{Adversarial Training}

Adversarial training includes retraining of the ML model using the adversarial examples along with the legitimate examples. This increases the robustness of the ML model against adversarial examples as the model is now able to learn a better distribution. Although adversarial retraining can help improve the robustness of the ML model, the ML model can still be compromised through adversarial examples generated through some other methods. The goal of adversarial training is to improve the generalization outside of the training manifold. Kos and Song \cite{kos2017delving} proposed using adversarial training for robustifying DRL algorithms. They retrain their agent on perturbations generated using FGSM and random noise and show performance retention against similar attacks. Furthermore, they observe that the retrained agent is also resilient against FGSM perturbations having magnitudes different than the one used for retraining.

Pattanaik et al.~\cite{pattanaik2018robust} also adopt adversarial training as a measure to make the algorithms robust against gradient-based attacks. They show its equivalence to Robust Control. They train the DRL model by using the adversarial samples generated from the gradient-based attacks. This helps the algorithm to model uncertainties in the system making them robust to similar adversarial attacks. They show that the addition of noise to the training samples while training increases the resilience of the DRL models against adversarial attacks. Han et al.~\cite{han2018reinforcement} also propose adversarial training as a method of robustifying the DRL algorithms against adversarial attacks. They show that this technique is effective when countering attacks, such as node-corruption and node-falsifying, in SDN.

Behzadan and Munir \cite{behzadan2017whatever} find the adversarially-trained policies to be more robust to test-time attacks. They investigate the robustness of DRL algorithms to both training and test-time attacks and find out that under the training time attack the DQN can learn and become robust by changing the policy. They propose that for an agent to recover from adversarial attacks, the number of the adversarial samples in the memory needs to reach a critical limit. In this way, when the agent samples a random batch from the memory, it can learn the perturbation statistics. They also compare the performance of $\epsilon$-greedy and parameter-space noise exploration methods in case of adversarial attacks. They show the $\epsilon$-greedy methods to be more robust to training-time attacks than the noisy exploration technique. They also find noisy exploration techniques to be able to recover faster from attacks when compared to the $\epsilon$-greedy methods.

Later on, Behzadan and Munir \cite{behzadan2018mitigation} compare the resilience to adversarial attacks of two DQNs: one based on $\epsilon$-greedy policy learning and another employed NoisyNets \cite{fortunato2017noisy} which is a parameter-space noise exploration technique. Their results show the NoisyNets to be more resilient to training-time attacks than that of the $\epsilon$-greedy policy. They argue that this resilience is due to the enhanced generalize-ability and reduced transferability in NoisyNets. They propose that by using parameter-space noise exploration, the DRL algorithms can be made robust to attack techniques like FGSM. Chen et al.~\cite{chen2018gradient} propose a gradient-based adversarial training technique. They use adversarial perturbations generated using their proposed attacking algorithm, i.e., CDG, for retraining the RL agent. This approach can achieve a precision of 93.89\% in detecting adversarial samples. They prove that adversarial training using only a single adversarial example, generated using CDG, can realize the generalized CDG-attack immunity of A3C pathfinding with high confidence. Behzadan and Hsu \cite{behzadan2019analysis} propose adversarially guided exploration (AGE) after considering the sample inefficiency of current adversarial training techniques. Their technique is based on a modified hybrid of the $\epsilon$-greedy algorithm and the Boltzmann exploration (exploring probabilistically relative to expected rewards). The new adversarial training procedure is tested on DQN trained for the CartPole environment with different perturbation probabilities. They show that for small perturbations probabilities, i.e., 0.2 and 0.4, the agent can recover from the attack while in the case of large probabilities such as 0.8 or 1 the agent is not able to recover. They compare the efficiency of their proposed technique with $\epsilon$-greedy and parameter-space noise exploration algorithms and prove its feasibility. 

Tan et al.~\cite{tan2020robustifying} argue that although the DRL algorithms used for decision and control tasks are vulnerable to adversarial attacks, little research has been done to make them robust. After showing the vulnerability of well-trained DRL agents to action space attacks, they use adversarial training to increase the robustness of the attacked model. Furthermore, a performance improvement of the adversarially trained agent over the normal agent in non-adversarial scenarios is also shown. Lee et al.~\cite{lee2020query} also show that the use of adversarial training with their proposed attack, decreases the attack success rate to half.

Vinitsky et al.~\cite{vinitsky2020robust} argue that existing literature on robust learning in DRL focuses on training a single RL agent against a single adversary and these systems are bound to fail in the case of a different adversary. They propose a population-based augmentation to the Robust RL formulation in which a population of adversaries is randomly initialized and samples are drawn uniformly from the population during training.

\subsection{Game-theoretic Approach}

Pinto et al.~\cite{pinto2017robust} propose \textit{robust adversarial reinforcement learning} ($RARL$) as a method of robust policy learning in the presence of an adversary. They formulate policy learning as a zero-sum minimax objective function to ensure robustness to differences in test and train conditions, even in the presence of an adversary. They use a self-proposed adversarial agent with a specially designed reward targeted at finding the state space trajectories that lead to the worst rewards. They call these trajectories hard examples. An adversary is introduced in the environment whose goal is to destabilize the RL agent. Abdullah et al.~\cite{abdullah2019wasserstein} propose a robust reinforcement learning using a novel min-max game with a Wasserstein constraint for a correct and convergent solver. This technique shows a significant increase in robustness in the case of both low and high-dimensional control tasks. They also discuss that by using their technique the DDPG algorithms are not able to achieve significant performance improvement in robustness, even in the case of Inverted Pendulum. While the other two DRL schemes, i.e., TRPO and PPO, demonstrate acceptable performance and hence are reported in their results.

Bravo and Mertikopoulos \cite{bravo2017robustness} examine a game approach where the players adjust their actions based on past payoff observations that are subject to adversarial perturbations. In the single-player case containing an agent trying to adapt to an arbitrarily changing environment, they show that irrespective of the level of noise in the player's observations, the stochastic dynamics under study leads to no regret almost surely. In the case of multiple players, they show that the dominated strategies become extinct and the strict Nash equilibrium is stochastically stable and attractive. Conversely, a stable or attractive state with better probability is the Nash equilibrium. Finally, they provide an averaging principle and show that in the case of 2-player zero-sum games with an interior equilibrium, the time averages converge to Nash equilibrium for any noise level. Ogunmolu et al.~\cite{ogunmolu2018minimax} propose an iterative minimax dynamic game framework that helps in designing robust policies in the presence of adversarial inputs. They also propose a method of quantifying the robustness capacity of a policy. They evaluate their proposed framework on a mecanum-wheeled robot. The goal of this agent is to find a locally robust optimal multistage policy that achieves a given goal-reaching task.

\subsection{Robust Learning}

Robust learning is a training mechanism to ensure robustness against training-time adversarial attacks. Behzadan and Munir \cite{behzadan2018mitigation} propose adding noise to the parameter state while training, this technique is found very effective in mitigating the effects of both training and test time attacks for both black-box and white-box settings. The results of the proposed method are tested on DQN trained for three Atari games, namely Enduro, Assault, and Blackout. In particular, the authors use FGSM for crafting adversarial samples. Then, they show the performance of the normal agents to deteriorate significantly, while the ones which were retrained using the parameter noise show great performance even in the presence of adversarial inputs. Mandlekar et al.~\cite{mandlekar2017adversarially} show superior resilience to adversarial attacks by introducing an \textit{adversarially robust policy learning} (ARPL) algorithm. This involves the use of adversarial examples during training to enable robust policy learning. They consider the addition of adversarial perturbations not only to the image space but to the whole state of the system which in their case also included the parameters like friction, mass, and inertia. They use the gradient-based FGSM technique for the generation of adversarial samples. They show in the case of agents that do not follow their learning technique the performance deteriorates drastically, while their agent can retain the training performance. It is important to note that the agent trained using the ARPL algorithm does not perform as well as the normal one in case of no perturbations.

Wang et al.~\cite{wang2018reinforcement} point out that the reward function is susceptible to 3 kinds of noise, namely inherent noise, application-specific noise, and adversarial noise. As a remedy, they propose a reward confusion matrix to generate rewards to help the RL agent to learn in cases of perturbed/noisy inputs. Such rewards are called to be \textit{Perturbed Rewards}. Using these perturbed rewards, they can develop an unbiased reward estimator aided robust RL framework. Their algorithm not only achieves higher expected rewards but also converges faster. They experiment with their technique extensively using several DRL algorithms which are trained for different classic Atari gaming environments. Their proposed technique can achieve 67.5\% and 46.7\% improvements in average reward when the error rate is 10\% and 30\%, respectively in the case of PPO. They discuss both the cases of the perturbations added to some samples and perturbations being added to all samples.\par

Policies that can retain the performance in non-stationary environments are also robust to adversarial attacks that involve adding noise to the state space. Smirnova et al.~\cite{smirnova2019distributionally} propose a distributionally robust policy iteration scheme to restrict the agent from learning sub-optimal policy while exploring in cases of high-dimensional state/action space. This induces a dynamic level of risk to stop the agent from taking sub-optimal actions. Their scheme is based on robust Bellman operators, which provide a lower-bound guarantee on the policy/state values. They also present a distributionally robust soft actor-critic based on mixed exploration, acting conservatively in the short-term and exploring optimistically in a long run leading to an optimal policy. The direct target in \cite{mandlekar2017adversarially} is adversarial robustness while in \cite{smirnova2019distributionally} the direct target is distributional robustness, hence, the target of adversarial robustness was achieved indirectly.

Tessler et al.~\cite{tessler2019action} propose \textit{probabilistic MDP} (PR-MDP) and \textit{noisy action robust MDP} (NR-MDP) as two new criteria for robustness. They modify the DDPG to form AR-DDPG for solving these MDPs. The proposed techniques are evaluated in various Mojuco environments and the results prove that the learning of action-robust policies can help in making the proposed algorithms secure and perform better even in the absence of these perturbations. The adversarial robustness was achieved here by making the agent ``\textit{action robust}".

Kumar et al.~\cite{kumar2019enhancing} present a technique to make the DRL algorithm learn in the presence of noisy rewards. The proposed scheme is based on using a neural network as a noise filter targeted at estimating the true reward of the environment. These estimates are then compared with the reward the agent gets at each state, to filter out the noisy samples. They show that beyond the perturbation probability of 0.5, their agent starts to learn based on adversarial samples rather than the normal ones.

Fisher et al.~\cite{fischer2019online} propose the idea of \textit{Robust Student DQN} (RS-DQN). They propose to split the standard DQN into two networks, namely a student (policy) network $S$ and a Q-network. This $S$ network is robustly trained and used for exploration while the Q-network is trained normally. This permits online robust training while keeping the competitive performance of the Q-networks. They show that in the case of no attacks, the DQN and RS-DQN show the same performance. While in the case of adversarial attacks, the DQN fails while RS-DQN remains robust. Furthermore, they show that RS-DQN when combined with state-of-the-art adversarial training provides resilience to strong adversarial attacks during training and evaluation. 

Pan et al.~\cite{pan2019risk} argue that training RL on physical hardware is dangerous due to exploration and can also be slow due to high sample complexity. This puts forward a need for a robust learning algorithm that can make sure that the policy performs well in catastrophic situations. They propose \textit{risk-averse robust adversarial reinforcement learning} (RARARL), using a risk-averse agent and a risk-seeking adversary. They point out that the RARL technique proposed by \cite{pinto2017robust} has no explicit modeling and optimization of risk as they only optimize the expected control objective. An ensemble of q-value networks is used to model risk as to the variance of value functions. Their technique is similar to Bootstrapped DQNs \cite{osband2016deep} proposed to assist exploration but in this case, the purpose of the ensemble is to estimate variance. They test their approach on a self-driving environment using the TORCS simulator and show that a risk-averse agent handles the risk better and leads to fewer crashes than a normal agent trained in a similar environment. The attacker and the victim agent are made to work independently in the same environment. This gives more options to the attacker to perturb the environment hence serving as a strong adversary.

The overfitting of RL policies to the training environments cause them to fail to generalize to safety-critical scenarios. Wang et al.~\cite{wang2020falsification} argue that the RARL technique proposed by \cite{pinto2017robust} requires handcrafting sophisticated reward signals which is a difficult task. They say that safety falsification methods can be used to find a set of initial conditions as well as an input sequence to make a system violate a given property formulated in temporal logic. They propose a falsification-based RARL (FRARL) technique for integrating temporal-logic falsification in adversarial learning to improve policy robustness. This removes the requirement of the construction of an extra reward function for the adversary. Their experiments show that the policies trained with FRARL generalize better and show less violation of the safety specifications in test scenarios when compared to techniques similar to RARL.

Lutjens et al.~\cite{lutjens2020certified} argue that adversarial detection (detecting adversarial samples using specialized models) can only detect a perturbed input but they cannot propose an alternate action in case of an attack. They leverage the research on certified adversarial robustness to develop an online certified defense for DRL algorithms called "Certified Adversarial Robustness for RL" (CARRL). CARRL involves computing lower bounds on the state-action pairs and choosing a robust action in case of an adversarial attack. They make their technique certifiable robust by using robust optimization to consider worst-case uncertainties and to provide certificates on solution quality. They show the effectiveness of their technique on a DQN trained for a collision-avoidance system and a classic control task (cartpole) and using targeted-FGSM as an adversary. Their experiments show that CARRL can (1) recover and avoid the obstacles in case of an adversarial attack on collision avoidance system; (2) recover and achieve a sufficient reward in cartpole environment. Furthermore, they prove that although their technique reduces the computational efficiency of the DQN, it increases the robustness.

Zhang et al.~\cite{zhang2020robust} point out that the robustness for continuous-action space DRL has not got any attention and existing approaches lack proper theoretical justification. They prove that the classification techniques like that of adversarial training prove to be inefficient for many RL problems. They develop a theoretically principled policy regularization and propose a state-adversarial Markov decision process (SA-MDP). Their technique improves robustness under strong white-box attacks on state observations, including the two new attacks that they have proposed: the robust SARSA attack (RS attack) and maximal action difference attack (MAD attack). Furthermore, they show performance improvement in non-adversarial scenarios. They assume the adversary to (i) be stationary, deterministic, and Markovian, i.e., the adversary does not change with time; (ii) have bounded adversary power, i.e., the adversary can only perturb a specific number of states.

Oikarinen et al.~\cite{oikarinen2020robust} propose a method of training DRL agents robust to $l_p$-bounded attacks. They termed their technique RADIAL-RL. Furthermore, they propose a new metric greedy worst-case reward (GWC) for evaluating the performance of DRL algorithms against adversarial attacks. They show their technique to out-perform the state-of-the-art robust learning techniques \cite{fischer2019online}, \cite{zhang2020robust} under PGD attack.

Zhang et al.~\cite{zhang2021robust} propose a technique to enhance the robustness of DRL agents against learned adversary attacks, i.e., attacks in which the adversary is continuously learning. They term their technique as \textit{alternating training with learned adversaries} (ATLA). Their technique involves the training of an adversarial agent online together with the victim agent using policy gradient following the optimal adversarial attack framework.

\subsection{Adversarial Detection}

Adversarial detection involves the detection of adversarial samples using a model specially trained to segregate the true samples from the adversarial ones. In this way, we can disregard the adversarial inputs without modifying the original model. Lin et al.~\cite{lin2017detecting} propose a method of protecting the DRL algorithms from adversarial attacks by leveraging an action-conditioned frame prediction module. By using this technique, they can detect the presence of adversarial attacks and make the model robust by using the predicted frame instead of the adversarial frame. They also compare their results with other ML defense approaches to show the effectiveness of this technique. The techniques used for adversarial example generation are FGSM, Carlini \& Wagner \cite{carlini2017towards}, and Basic Iterative Method \cite{kurakin2016adversarial}. The present results indicate that their proposed technique can detect adversarial attacks with accuracy from 60\% to 100\%. 

Havens et al.~\cite{havens2018online} detect the presence of adversarial attacks via a supervisory agent by learning separate sub-policies using the \textit{meta-learned advantage hierarchy} (MLAH) framework. Because this technique can handle the attacks in the decision space, it can mitigate the learned bias introduced by the adversary. They consider a policy learning problem that is being attacked at specific periods. The goal of the adversary is the corruption of state-space while the agent is been trained. They assume that while training, the agent learns sub-policies before learning the ultimate policy. Thus, the supervisory agent can detect the presence of the adversarial examples due to them being unexpected. They use a self-defined adversarial agent having the ability to perturb the states before they reach the agent for training. The perturbations generated by this agent are bounded by $l_\infty$-norm.

Xiang et al.~\cite{xiang2018pca} propose an advanced Q-learning algorithm for automatic path-finding in robots, that is robust to adversarial attacks by detecting the adversarial inputs. Specifically, they propose a model to predict the adversarial inputs based on a calculation determined by 5 factors: energy point gravitation, key point gravitation, path gravitation, included angle, and the placid point. The weights for these 5 factors are calculated based on the principal component analysis (PCA). Using these factors, they train a model able to achieve a precision of 70\% in segregating adversarial inputs from the normal ones.

Gallego et al.~\cite{gallego2019reinforcement} introduce \textit{threatened Markov decision processes} (TMDPs), a variant of MDP. This framework supports the decision-making process in the DRL setting against adversaries that affect the reward generating process. They propose a level-k thinking scheme resulting in a new framework for dealing with TMDPs. They show that while a normal Q-learning algorithm is exploited by an adversary, a level-2 learner can approximately estimate the adversarial behavior and achieve a positive reward. Integrating DQNs to TMDPs is discussed as a future research path.\par

\subsection{Defensive Distillation}

Papernot et al.~\cite{papernot2016distillation} propose the idea of using defensive distillation to deal with adversarial attacks on ML schemes. In this technique, one model is trained to predict the output probabilities of another model that was trained with an emphasis on accuracy. Using this technique, DL models can be made less susceptible to exploitation by adding flexibility to an algorithm's classification process using adversarial training. Carlini and Wagner \cite{carlini2016defensive} show that defensive distillation gives a false sense of robustness against adversarial examples. Rusu et al.~\cite{rusu2015policy} present a method of extracting the policy of a dense network to train another comparatively less dense network. This new network can take expert-level decisions while being smaller in size. This method can also be used to merge multiple task-specific policies into a single policy. They show that the distilled agents which were 4 times smaller than DQNs were able to achieve better performance than DQN. They also show that the agents having 25 times fewer parameters than DQN were able to achieve a performance of 84\% as compared to 100\% of the DQN. Such networks are proved to be more stable and robust to adversarial noise and attacks, as they have fewer parameters than their denser counterparts and hence decreasing the count of attack-able parameters.\par

Recently, Czarnecki et al.~\cite{pmlr-v89-czarnecki19a} analyzed empirically and theoretically each variant of distillation and reported the strengths and weaknesses of each variant. Furthermore, they propose \textit{expected entropy regularized distillation} which makes the training much faster while guaranteeing convergence. This technique can be used in making the DRL models robust to adversarial attacks by leveraging learning information from a complex model into a simpler one. Hence, making the models robust to adversarial attacks. However, as discussed by Carlini and Wagner~\cite{carlini2016defensive}, using this technique alone may not be effective. It needs to be combined with other approaches, like adversarial training, adversarial detection, etc., to be successful.\par

Qu et al.~\cite{qu2020defending} propose robust policy distillation, i.e., a policy distillation paradigm capable of achieving an adversarially robust student policy without relying on any adversarial example during student policy training. They propose a policy distillation loss consisting of a prescription gap maximization (PGM) loss and a Jacobian regularization (JR) loss. They perform a theoretical analysis and show that their proposed mechanism ensures the learning of robust policies during the distillation process. They show their technique to outperform the one proposed in \cite{zhang2020robust}.\par

\begin{table*}[!ht]
\centering
\caption{\MakeUppercase{Summary of defenses against adversarial attacks on DRL}}
\label{tab:defenses}
\resizebox{\textwidth}{!}{%
\begin{tabular}{ccccc}

\hline
\multicolumn{5}{|c|}{\textbf{Adversarial Training}} \\ \hline
\multicolumn{1}{|c|}{\multirow{2}{*}{\textbf{Paper}}} & \multicolumn{1}{c|}{\multirow{2}{*}{\textbf{Proposed Techniques}}} & \multicolumn{2}{c|}{\textbf{Setup}} & \multicolumn{1}{c|}{\multirow{2}{*}{\textbf{Effective Against}}} \\ \cline{3-4}
\multicolumn{1}{|c|}{} & \multicolumn{1}{c|}{} & \multicolumn{1}{c|}{\textbf{Algorithm}} & \multicolumn{1}{c|}{\textbf{Environment}} & \multicolumn{1}{c|}{} \\ \hline
\multicolumn{1}{|c|}{Kos and Song~\cite{kos2017delving}} & \multicolumn{1}{c|}{Adversarial Training using Random Noise and FGSM} & \multicolumn{1}{c|}{A3C} & \multicolumn{1}{c|}{Pong} & \multicolumn{1}{c|}{Random Noise FGSM Attacks} \\ \hline
\multicolumn{1}{|c|}{Pattanaik et al.~\cite{pattanaik2018robust}} & \multicolumn{1}{c|}{\begin{tabular}[c]{@{}c@{}}Adversarial Training using Noise Gradient-Based\\ Attacks\end{tabular}} & \multicolumn{1}{c|}{DDQN, DDPG} & \multicolumn{1}{c|}{\begin{tabular}[c]{@{}c@{}}Cartpole, Mountain Car,\\ Hopper, Half Cheetah\end{tabular}} & \multicolumn{1}{c|}{Noise Gradient Based Attacks} \\ \hline
\multicolumn{1}{|c|}{Han et al.~\cite{han2018reinforcement}} & \multicolumn{1}{c|}{Adversarial Training using corrupted nodes in SDN} & \multicolumn{1}{c|}{DDQN, A3C} & \multicolumn{1}{c|}{SDN} & \multicolumn{1}{c|}{Node Corruption Falsifying Attacks} \\ \hline
\multicolumn{1}{|c|}{Behzadan and Munir~\cite{behzadan2017whatever}} & \multicolumn{1}{c|}{Adversarial Training using Perturbed States} & \multicolumn{1}{c|}{DQN} & \multicolumn{1}{c|}{Breakout, Pong} & \multicolumn{1}{c|}{Attacks Perturbing a considerable no. of States} \\ \hline
\multicolumn{1}{|c|}{Behzadan and Munir~\cite{behzadan2018mitigation}} & \multicolumn{1}{c|}{Noisy Exploration} & \multicolumn{1}{c|}{DQN} & \multicolumn{1}{c|}{\begin{tabular}[c]{@{}c@{}}Enduro, Assault, \\ Blackout\end{tabular}} & \multicolumn{1}{c|}{State Perturbation Attacks} \\ \hline
\multicolumn{1}{|c|}{Chen et al.~\cite{chen2018gradient}} & \multicolumn{1}{c|}{Gradient Band-Based Adversarial Training} & \multicolumn{1}{c|}{A3C} & \multicolumn{1}{c|}{Pathfinding} & \multicolumn{1}{c|}{Gradient Band Based Adversarial Attacks} \\ \hline
\multicolumn{1}{|c|}{Behzadan and Hsu~\cite{behzadan2019analysis}} & \multicolumn{1}{c|}{Adversarially Guided Exploration (AGE)} & \multicolumn{1}{c|}{DQN} & \multicolumn{1}{c|}{Cartpole} & \multicolumn{1}{c|}{Limited Attack Samples} \\ \hline
\multicolumn{1}{|c|}{Tan et al.~\cite{tan2020robustifying}} & \multicolumn{1}{c|}{Adversarial Training} & \multicolumn{1}{c|}{PPO} & \multicolumn{1}{c|}{Lunar Lander} & \multicolumn{1}{c|}{Action space Attacks} \\ \hline
\multicolumn{1}{|c|}{Lee et al.~\cite{lee2020query}} & \multicolumn{1}{c|}{Adversarial Training} & \multicolumn{1}{c|}{PPO} & \multicolumn{1}{c|}{point goal, car goal \cite{ray2019benchmarking}} & \multicolumn{1}{c|}{Action space Attacks} \\ \hline
\multicolumn{1}{|c|}{Vinitsky et al.~\cite{vinitsky2020robust}} & \multicolumn{1}{c|}{Adversarial Training using Populations} & \multicolumn{1}{c|}{PPO} & \multicolumn{1}{c|}{Hopper, Ant, Half-cheetah} & \multicolumn{1}{c|}{Generic Adversarial Attacks} \\ \hline
 & & & & \\ \hline

\multicolumn{5}{|c|}{\textbf{Robust Learning}} \\ \hline
\multicolumn{1}{|c|}{\multirow{2}{*}{\textbf{Paper}}} & \multicolumn{1}{c|}{\multirow{2}{*}{\textbf{Proposed Techniques}}} & \multicolumn{2}{c|}{\textbf{Setup}} & \multicolumn{1}{c|}{\multirow{2}{*}{\textbf{Effective Against}}} \\ \cline{3-4}
\multicolumn{1}{|c|}{} & \multicolumn{1}{c|}{} & \multicolumn{1}{c|}{\textbf{Algorithm}} & \multicolumn{1}{c|}{\textbf{Environment}} & \multicolumn{1}{c|}{} \\ \hline
\multicolumn{1}{|c|}{Mandlekar et al.~\cite{mandlekar2017adversarially}} & \multicolumn{1}{c|}{\begin{tabular}[c]{@{}c@{}}Adversarially Robust Policy Learning\\ (ARPL)\end{tabular}} & \multicolumn{1}{c|}{TRPO} & \multicolumn{1}{c|}{\begin{tabular}[c]{@{}c@{}}Inverted Pendulum, Half-\\ Cheetah, Hopper, Walker\end{tabular}} & \multicolumn{1}{c|}{State Perturbation Attacks} \\ \hline
\multicolumn{1}{|c|}{Smirnova et al.~\cite{smirnova2019distributionally}} & \multicolumn{1}{c|}{Distributionally Robust Policy Iteration} & \multicolumn{1}{c|}{self} & \multicolumn{1}{c|}{Hopper, Walker2D} & \multicolumn{1}{c|}{Attacks Targeting the Policy} \\ \hline
\multicolumn{1}{|c|}{Tessler et al.~\cite{tessler2019action}} & \multicolumn{1}{c|}{PR-MDPs, NR-MDPs} & \multicolumn{1}{c|}{self} & \multicolumn{1}{c|}{\begin{tabular}[c]{@{}c@{}}Hopper, Walker2d,\\ Humanoid, Inverted-Pendulum\end{tabular}} & \multicolumn{1}{c|}{Generic Adversarial Attacks} \\ \hline
\multicolumn{1}{|c|}{Kumar et al.~\cite{kumar2019enhancing}} & \multicolumn{1}{c|}{Noise Filter} & \multicolumn{1}{c|}{DQN, DDQN} & \multicolumn{1}{c|}{cartpole} & \multicolumn{1}{c|}{Attacks Perturbing the Rewards} \\ \hline
\multicolumn{1}{|c|}{Fisher et al.~\cite{fischer2019online}} & \multicolumn{1}{c|}{Robust Student DQN (RS-DQN)} & \multicolumn{1}{c|}{self} & \multicolumn{1}{c|}{\begin{tabular}[c]{@{}c@{}}Freeway, BankHeist,\\ Pong, boxing, road-runner\end{tabular}} & \multicolumn{1}{c|}{Generic Adversarial Attacks} \\ \hline
\multicolumn{1}{|c|}{Lutjens et al.~\cite{lutjens2020certified}} & \multicolumn{1}{c|}{Certified Adversarial Robustness for RL (CARRL)} & \multicolumn{1}{c|}{DQN} & \multicolumn{1}{c|}{collision avoidance, cartpole} & \multicolumn{1}{c|}{Generic Adversarial Attacks} \\ \hline
\multicolumn{1}{|c|}{Pan et al.~\cite{pan2019risk}} & \multicolumn{1}{c|}{\begin{tabular}[c]{@{}c@{}}Risk-Averse Robust Adversarial\\ Reinforcement Learning (RARARL)\end{tabular}} & \multicolumn{1}{c|}{} & \multicolumn{1}{c|}{TORCS} & \multicolumn{1}{c|}{Generic Adversarial Attacks} \\ \hline
\multicolumn{1}{|c|}{Wang et al.~\cite{wang2020falsification}} & \multicolumn{1}{c|}{Falsification-based RARL (FRARL)} & \multicolumn{1}{c|}{PPO} & \multicolumn{1}{c|}{\begin{tabular}[c]{@{}c@{}}brake assistance system,\\ adaptive cruise control system\end{tabular}} & \multicolumn{1}{c|}{Generic Adversarial Attacks} \\ \hline
\multicolumn{1}{|c|}{Zhang et al.~\cite{zhang2020robust}} & \multicolumn{1}{c|}{\begin{tabular}[c]{@{}c@{}}State-Adversarial Markov Decision\\ Process (SA-MDP)\end{tabular}} & \multicolumn{1}{c|}{\begin{tabular}[c]{@{}c@{}}DQN, PPO, \\ DDPG\end{tabular}} & \multicolumn{1}{c|}{\begin{tabular}[c]{@{}c@{}}Walker, Hopper, Humanoid, Ant,\\ Inverted Pendulum, Reacher,\\ Road Runner, BankHeist, pong, \\ Acrobot, and Freeway\end{tabular}} & \multicolumn{1}{c|}{Generic Adversarial Attacks} \\ \hline
\multicolumn{1}{|c|}{Oikarinen et al.~\cite{oikarinen2020robust}} & \multicolumn{1}{c|}{RADIAL-RL} & \multicolumn{1}{c|}{DQN, A3C} & \multicolumn{1}{c|}{\begin{tabular}[c]{@{}c@{}}Pong, Freeway, bankheist,\\ road runner\end{tabular}} & \multicolumn{1}{c|}{Generic Adversarial Attacks} \\ \hline
\multicolumn{1}{|c|}{Zhang et al.~\cite{zhang2021robust}} & \multicolumn{1}{c|}{\begin{tabular}[c]{@{}c@{}}Alternating Training with Learnable\\ Adversaries (ATLA)\end{tabular}} & \multicolumn{1}{c|}{PPO} & \multicolumn{1}{c|}{\begin{tabular}[c]{@{}c@{}}Hopper, Walker2D, Ant,\\ Half-Cheetah\end{tabular}} & \multicolumn{1}{c|}{Generic Adversarial Attacks} \\ \hline
\multicolumn{1}{|c|}{Wang et al.~\cite{wang2018reinforcement}} & \multicolumn{1}{c|}{Reward Confusion Matrix} & \multicolumn{1}{c|}{\begin{tabular}[c]{@{}c@{}}Q-Learning, \\ CEM, SARSA,\\ DQN, PPO, NAF,\\ Dueling DQN, DDPG\\ \end{tabular}} & \multicolumn{1}{c|}{\begin{tabular}[c]{@{}c@{}}CartPole, Pendulum,\\ AirRaid, Alien, Carnival,\\ MsPacman, Pong,\\ Phoenix, Seaquest\end{tabular}} & \multicolumn{1}{c|}{Attacks Perturbing the Rewards} \\ \hline
 & & & & \\ \hline
 
\multicolumn{5}{|c|}{\textbf{Adversarial Detection}} \\ \hline
\multicolumn{1}{|c|}{\multirow{2}{*}{\textbf{Paper}}} & \multicolumn{1}{c|}{\multirow{2}{*}{\textbf{Proposed Techniques}}} & \multicolumn{2}{c|}{\textbf{Setup}} & \multicolumn{1}{c|}{\multirow{2}{*}{\textbf{Effective Against}}} \\ \cline{3-4}
\multicolumn{1}{|c|}{} & \multicolumn{1}{c|}{} & \multicolumn{1}{c|}{\textbf{Algorithm}} & \multicolumn{1}{c|}{\textbf{Environment}} & \multicolumn{1}{c|}{} \\ \hline
\multicolumn{1}{|c|}{Lin et al.~\cite{lin2017detecting}} & \multicolumn{1}{c|}{\begin{tabular}[c]{@{}c@{}}Action-conditioned Frame Prediction\\ Module\end{tabular}} & \multicolumn{1}{c|}{DQN} & \multicolumn{1}{c|}{\begin{tabular}[c]{@{}c@{}}Pong, Freeway, Sea-quest,\\ Chopper-Command, Ms-Pacman\end{tabular}} & \multicolumn{1}{c|}{Attacks Perturbing the States} \\ \hline
\multicolumn{1}{|c|}{Havens et al.~\cite{havens2018online}} & \multicolumn{1}{c|}{\begin{tabular}[c]{@{}c@{}}Meta-learned Advantage Hierarchy\\ (MLAH)\end{tabular}} & \multicolumn{1}{c|}{self} & \multicolumn{1}{c|}{\begin{tabular}[c]{@{}c@{}}InvertedPendulum-v2,\\ MountainCarContinuous-v0,\\ Hopper-v2\end{tabular}} & \multicolumn{1}{c|}{Training-Time Poisoning Attacks} \\ \hline
\multicolumn{1}{|c|}{Xiang et al.~\cite{xiang2018pca}} & \multicolumn{1}{c|}{PCA for Adversarial Detection} & \multicolumn{1}{c|}{Q-learning} & \multicolumn{1}{c|}{Pathfinding} & \multicolumn{1}{c|}{Attacks Perturbing the States} \\ \hline
\multicolumn{1}{|c|}{Gallego et al.~\cite{gallego2019reinforcement}} & \multicolumn{1}{c|}{Threatened Markov Decision Processes (TMDPs)} & \multicolumn{1}{c|}{self} & \multicolumn{1}{c|}{Chicken game} & \multicolumn{1}{c|}{Attacks Affecting Reward Generation} \\ \hline
 & & & & \\ \hline
 
\multicolumn{5}{|c|}{\textbf{Defensive Distillation}} \\ \hline
\multicolumn{1}{|c|}{\multirow{2}{*}{\textbf{Paper}}} & \multicolumn{1}{c|}{\multirow{2}{*}{\textbf{Proposed Techniques}}} & \multicolumn{2}{c|}{\textbf{Setup}} & \multicolumn{1}{c|}{\multirow{2}{*}{\textbf{Effective Against}}} \\ \cline{3-4}
\multicolumn{1}{|c|}{} & \multicolumn{1}{c|}{} & \multicolumn{1}{c|}{\textbf{Algorithm}} & \multicolumn{1}{c|}{\textbf{Environment}} & \multicolumn{1}{c|}{} \\ \hline
\multicolumn{1}{|c|}{Rusu et al.~\cite{rusu2015policy}} & \multicolumn{1}{c|}{Defensive Distillation} & \multicolumn{1}{c|}{DQN} & \multicolumn{1}{c|}{\begin{tabular}[c]{@{}c@{}}pong, space-invaders, Breakout,\\ Freeway, Q-bert,Beamrider,\\ Enduro, MS. Pacman, Seaquest,\\ riverraid\end{tabular}} & \multicolumn{1}{c|}{Generic Adversarial Attacks} \\ \hline
\multicolumn{1}{|c|}{Qu et al.~\cite{qu2020defending}} & \multicolumn{1}{c|}{Defensive Distillation} & \multicolumn{1}{c|}{\begin{tabular}[c]{@{}c@{}}DDQN, Rainbow\\ DQN\end{tabular}} & \multicolumn{1}{c|}{\begin{tabular}[c]{@{}c@{}}Freeway, bankheist, pong,\\ boxing, roadrunner\end{tabular}} & \multicolumn{1}{c|}{Generic Adversarial Attacks} \\ \hline
 & & & & \\ \hline
 
\multicolumn{5}{|c|}{\textbf{Game Theoretic Approach}} \\ \hline
\multicolumn{1}{|c|}{\multirow{2}{*}{\textbf{Paper}}} & \multicolumn{1}{c|}{\multirow{2}{*}{\textbf{Proposed Techniques}}} & \multicolumn{2}{c|}{\textbf{Setup}} & \multicolumn{1}{c|}{\multirow{2}{*}{\textbf{Effective Against}}} \\ \cline{3-4}
\multicolumn{1}{|c|}{} & \multicolumn{1}{c|}{} & \multicolumn{1}{c|}{\textbf{Algorithm}} & \multicolumn{1}{c|}{\textbf{Environment}} & \multicolumn{1}{c|}{} \\ \hline
\multicolumn{1}{|c|}{Pinto et al.~\cite{pinto2017robust}} & \multicolumn{1}{c|}{\begin{tabular}[c]{@{}c@{}}Robust Adversarial Reinforcement\\ Learning (RARL)\end{tabular}} & \multicolumn{1}{c|}{TRPO} & \multicolumn{1}{c|}{\begin{tabular}[c]{@{}c@{}}InvertedPendulum, Half-\\ Cheetah, Hopper,\\ Swimmer, Walker2D\end{tabular}} & \multicolumn{1}{c|}{Attacks Targeting the Performance} \\ \hline
\multicolumn{1}{|c|}{Abdullah et al.~\cite{abdullah2019wasserstein}} & \multicolumn{1}{c|}{\begin{tabular}[c]{@{}c@{}}Wasserstein Robust Reinforcement\\ Learning ($WR^2L$)\end{tabular}} & \multicolumn{1}{c|}{\begin{tabular}[c]{@{}c@{}}DDPG, TRPO, \\ PPO\end{tabular}} & \multicolumn{1}{c|}{\begin{tabular}[c]{@{}c@{}}CartPole, Hopper,\\ Halfcheetah, Walker2D\end{tabular}} & \multicolumn{1}{c|}{Generic Adversarial Attacks} \\ \hline
\multicolumn{1}{|c|}{Bravo and Mertikopoulos~\cite{bravo2017robustness}} & \multicolumn{1}{c|}{Game-Theoretic Approach} & \multicolumn{1}{c|}{-} & \multicolumn{1}{c|}{-} & \multicolumn{1}{c|}{Noise Based Attacks} \\ \hline
\multicolumn{1}{|c|}{Ogunmolu et al.~\cite{ogunmolu2018minimax}} & \multicolumn{1}{c|}{Game-Theoretic Approach} & \multicolumn{1}{c|}{self} & \multicolumn{1}{c|}{Goal Reaching Task} & \multicolumn{1}{c|}{Attacks Targeting the Policy} \\ \hline
 & & & & \\ \hline
 
\multicolumn{5}{|c|}{\textbf{Others}} \\ \hline
\multicolumn{1}{|c|}{\multirow{2}{*}{\textbf{Paper}}} & \multicolumn{1}{c|}{\multirow{2}{*}{\textbf{Proposed Techniques}}} & \multicolumn{2}{c|}{\textbf{Setup}} & \multicolumn{1}{c|}{\multirow{2}{*}{\textbf{Effective Against}}} \\ \cline{3-4}
\multicolumn{1}{|c|}{} & \multicolumn{1}{c|}{} & \multicolumn{1}{c|}{\textbf{Algorithm}} & \multicolumn{1}{c|}{\textbf{Environment}} & \multicolumn{1}{c|}{} \\ \hline
\multicolumn{1}{|c|}{Behzadan and Munir~\cite{behzadan2018adversarial}} & \multicolumn{1}{c|}{Benchmarking} & \multicolumn{1}{c|}{DDPG} & \multicolumn{1}{c|}{Torcs collision avoidance} & \multicolumn{1}{c|}{Generic Adversarial Attacks} \\ \hline
\multicolumn{1}{|c|}{Behzadan and Hsu~\cite{behzadan2019sequential}} & \multicolumn{1}{c|}{Water Marking} & \multicolumn{1}{c|}{DQN} & \multicolumn{1}{c|}{cartpole} & \multicolumn{1}{c|}{Model Extraction Attacks} \\ \hline
\end{tabular}%
}
\end{table*}

\subsection{Discussion}
We discuss the state-of-the-art defenses in this section by categorizing them into adversarial training, robust learning, adversarial detection, defensive distillation, and game-theoretic approaches. It can be seen that most of these techniques are only effective against the specified type of adversarial attacks and do not provide any guarantees against other types of attacks. Most of these techniques focus on making the DRL agent learn a robust policy by use of different mechanisms like training using adversarial examples, simulating min-max games with adversaries, using robust alternatives of MDPs, etc. These techniques are more practical than the ones involving adversarial detection. Due to the advent of new attacking strategies by the day, one can never be sure that a detection mechanism will be able to detect the attack. Furthermore, it is worth noting that there are very few defenses for DRL algorithms that do not involve images as the observations.

Table \ref{tab:defenses} summarizes key information of the proposed defenses for DRL algorithms.

\section{Metrics, tools, and platforms for benchmarking DRL}
\label{sec:bench}

As we have previously discussed, DRL is different from other ML schemes, and only reporting the accuracy is not sufficient to cover security aspects of the DRL schemes. In particular, we need to consider the temporal domain aspect of the DRL while designing the DRL-based attack or defense. Benchmarking the DRL performance in attacks and defenses is very important. The need for an applicable solution to evaluate the robustness and resilience of DRL policies is not fulfilled by the current literature. There is also a need for a quantitative approach to measure and benchmark the resilience and robustness of DRL policies in a reusable and generalizable manner.

There are few benchmarks proposed, but they are not sufficient to cover the security aspects needed to measure the robustness and resilience of DRL algorithms. The few proposed approaches are discussed in this section. Behzadan and Hsu \cite{behzadan2019rl} introduce the terms of \textit{adversarial budget} and \textit{adversarial regret} as a measure to quantify the robustness and resilience of DRL algorithms. Adversarial budget is defined as the maximum number of features that can be perturbed in the observation, and the probability of perturbing each observation. The adversarial regret is the difference between the reward obtained by the unperturbed agent and the reward obtained by the perturbed agent after an episode. Based on these two terms, Behzadan and Hsu \cite{behzadan2019rl} define test-time resilience and test-time robustness.

\subsection{Test-time Resilience and Robustness}
\textit{Test-time resilience} is described as the minimum number of perturbations required to incur the maximum reduction to return at time $t$, while \textit{test-time robustness} is described as the maximum achievable adversarial regret.

The following procedure is proposed to measure test-time resilience for DRL algorithms:
\begin{itemize}
 \item Approximate the state-action value function using policy imitation in case it is not already given.
 \item Report the optimal adversarial return and maximum adversarial regret by training the adversarial agent against the target's policy.
 \item Apply the obtained adversarial policy to the target for several episodes while recording the return for each episode.
 \item Report the average adversarial return over these episodes as the mean test-time resilience of the target policy.
\end{itemize}

The method of measuring the test-time robustness is the same as test-time resilience. The only difference is that in the test-time case we measure the average adversarial regret in place of the average adversarial reward.

Behzadan and Munir \cite{behzadan2018adversarial} propose a novel framework for benchmarking the behavior of DRL-based collision avoidance mechanisms under the worst-case scenario of dealing with an adversarial agent which is trained to drive the system into unsafe states. They prove the practical applicability of the technique by comparing the reliability of two collision avoidance systems against intentional collision attempts. More recently, Behzadan and Hsu \cite{behzadan2019sequential} have presented a technique for watermarking DRL policies for robustness against model extraction attacks. This involves the integration of a unique response to a specific sequence of states while keeping its impact on performance minimum hence saving from the unauthorized replication of policies. It is shown that unwatermarked policies are not able to follow the identified trajectory.

\subsection{Metrics for Attack Performance}

Kiourti et al.~\cite{kiourti2019trojdrl}, introduce three metrics for measuring the performance of the DRL attacks, namely (1) performance gap, (2) percentage of target action, and (3) time to failure. As the name suggests, the performance gap is the difference between the performance of the normal and the victim model. For the second metric (percentage of target action), they measure the number of times the adversarial/targeted action is performed by the victim policy. The third metric (time to failure) is the number of consecutive states that need to be perturbed to trigger a complete failure of the model. 

As observed, these proposed measurement techniques can only cover a part of the DRL algorithms, and hence are not sufficient for measuring the performance of the DRL algorithms under the wide range of adversarial attacks and defenses. There is, therefore, a need for the development of benchmarks that can be used as standards for DRL algorithms as a measure of their resilience and robustness to adversarial attacks.

\subsection{Attacking DRL: Tools and Platforms}

DRL can be implemented using several available toolkits or by using a combination of these toolkits. Some of the ways to implement DRL are:

\begin{itemize}
 \item OpenAI Gym \cite{1606.01540} is a toolkit for testing the RL algorithms which provides with multiple gaming environments like pong, space-invaders, and lunar-lander. This toolkit is combined with Tensorflow \cite{tensorflow2015-whitepaper} to test the DRL algorithms. The DNN part can be implemented on the later one and then the choice of actions based on the states is done by the neural network.

 \item OpenAI Baselines \cite{baselines} provide a set of high-quality implementations of RL algorithms.
 
 \item RLCoach \cite{caspi_itai_2017_1134899} provides with integrated mechanisms of implementing the DNN and testing DRL algorithms.
 
 \item Horizon \cite{gauci2018horizon} is an open-source project (now known as ReAgent \cite{reagent}) that also provides integrated mechanisms for testing multiple DRL algorithms.
 
 \item Ns3-gym platform \cite{ns3gym} provides with network environments to test RL algorithms. This again can be combined with Tensorflow \cite{tensorflow2015-whitepaper} to test DRL algorithms.
\end{itemize}

All of these toolkits can be further combined with the toolkits available for attacking DL \cite{papernot2018cleverhans} and DRL \cite{behzadan2017whatever} to test different attacks and defenses on DRL algorithms in simulated environments.

\section{Open issues and research challenges}
\label{sec:open}

We identify the following major open issues and research challenges in DRL techniques. At the end of this section, we have also provided a roadmap to secure and robustify DRL.

\subsection{Universally Robust Algorithms}
Despite the presence of the various defenses that have been proposed, the security of DRL algorithms remains an open challenge. The proposed defenses are only able to defend from attacks they are designed for. Hence, they are still vulnerable to attacks led by proactive adversaries. Moosavi-Dezfooli et al.~\cite{moosavi2017analysis} point out that no matter how many adversarial examples are added to the training data, there are new adversarial examples that can be generated to cheat those newly trained networks. Moreover, if the adversary is only targeting confidence levels then we may never be able to detect the attack until the adversary uses his created deficiency for his benefit. We may not be even able to trace the attacks as shown by Clark et al.~\cite{clark2018malicious}. Thus, methods to make the DRL algorithms more robust are an urgent need. 

\subsection{Multitask Learning}
One of the major challenges for DRL is learning to do multiple tasks at a single time. It requires a lot of samples for this. Currently, proposed DRL algorithms can only learn to perform one task perfectly. They can be trained to play multiple games (like Cartpole, Inverted Pendulum, etc.), but they need to be trained from scratch for each game. The algorithms are expected to be scalable and be more generalizable so that their learning can be transferred from one game to another. Multi-task learning can help in making robust models that can grip the true essence of the tasks and hence become difficult to be fooled.

\subsection{Metrics for Robustness and Resilience}
We need to study why vulnerabilities exist in DRL models and how we can mitigate them and train robust models. A major reason for the existence of these vulnerabilities is the use of DRL models without the proper knowledge of the domain. There is a need to properly define the benchmarks of DRL in terms of the robustness of DRL against adversarial attacks. Behzadan and Hsu \cite{behzadan2019rl} have proposed techniques for quantifying the robustness and resilience of the RL algorithms. Some benchmarks are also proposed by Kiourti et al.~\cite{kiourti2019trojdrl} but as previously discussed these benchmarks are inadequate to measure the robustness and resilience of an algorithm even though they can be used as stepping stones to lead to a final goal. 

\subsection{System Design and Transferability}
System design remains an open challenge for the case of DRL. There is a need to define standards for system design for DRL problems as in this case the learning process is not supervised. So, the agent may not focus on the features that it needs to learn. This can introduce the error by mistake of the intermediary and also even induce his behavior on the model. We need to have proper standards for designing the reward functions. The system design needs to be robust and resilient to adversarial attacks.

\subsection{Ensemble of Defenses}
Various ensemble defenses have been proposed for the case of DL. However, they may not be appropriate to apply in the case of DRL as it can lead to an exponential increase in the complexity of the model which results in a significant decrease in performance. In the case of DRL, the model is making a real-time prediction, so a small reduction in the computation capabilities may cause a great loss to the agent. This remains a challenge to defend DRL models using an ensemble with a minimum loss of computations.

\subsection{Model Privacy}
Privacy has become a leading issue these days and model-extraction attacks pose a serious threat to the integrity of the learned models through illegal duplication. A mitigation for this, suggested by Behzadan and Hsu \cite{behzadan2019adversarial}, is to increase the cost of such attacks or to watermark the policies. We may experience some randomness in the agent to save from such attacks but that will incur an unacceptable loss of decreased performance. Developing techniques that can incur constrained randomization in the model to save from such attacks is a promising field of research.

\subsection{Explainable and Transparent DRL} For deploying AI systems in real-world scenarios, trust is a key component. The developer needs to be confident of the employed model's decisions. Transparency ensures that the model is fair and ethical while explainability helps to explain and justify the model's decisions. There are a few articles that discuss explainability for DRL \cite{puiutta2020explainable}, \cite{heuillet2021explainability}. Current techniques for explaining DRL do not specifically focus on targeting specific audiences, i.e., tester, developer, and the general public, and there is a need for the development of such techniques \cite{heuillet2021explainability}. Only through their development, we might be able to make DRL responsible, trustworthy, and applicable in critical applications.

\subsection{Transfer Learning for DRL}

Transfer learning \cite{tan2018survey} is a field of ML that does not require the training data and test data to be i.i.d.. The model is not needed to be trained from scratch in case of a different domain. Hence significantly reducing the demand of training data and training time of the target model. The research on transfer learning in the context of DRL has been limited and there are a few papers that test and discuss transfer learning for only some specific DRL algorithms \cite{zhu2020transfer}. Transfer learning can help with saving training time for these models; hence, making DRL more applicable to real-world scenarios.

\subsection{Roadmap Towards Secure and Robust DRL}
The ultimate goal of research in Artificial Intelligence (AI) is to develop Artificial General Intelligence (AGI) which can perform similar activities as humans in a more efficient manner. In addition to the algorithm being able to learn the task at hand efficiently, it also has to be computationally efficient while being robust to adversarial attacks. Furthermore, algorithms need to be sample-efficient to be able to learn quickly in real environments as there might not always be enough time to wait for the algorithm to converge. Meeting all these requirements at the same time is a challenging task and one has to strike intelligent tradeoffs between them.

In this regard, the first task to achieve is the development of sample-efficient and inherently robust DRL algorithms. The second task is the development of explainability techniques which can explain the behavior of these algorithms in accordance with human perception. The final task will be the development of metrics that can be used to quantify the robustness and resilience of these DRL algorithms. Based on these metrics and the sample and computational efficiency, one can then choose the most suitable algorithm for the task at hand.

\section{Conclusions}
\label{sec:con}
The broadening applicability of Deep Reinforcement Learning (DRL) in the real world has directed our concern to the security of these algorithms against adversarial attacks. This paper has provided a comprehensive survey of the latest techniques proposed for attacking DRL algorithms and the defenses proposed for defending against these attacks. We have also discussed the open research issues and provided the list of available benchmarks for measuring the resilience and robustness of DRL algorithms.

\section*{Acknowledgment}
This publication was made possible by NPRP grant \# [13S-0206-200273] from the Qatar National Research Fund (a member of Qatar Foundation). The statements made herein are solely the responsibility of the authors.

\bibliographystyle{IEEEtran}
\balance
\bibliography{references}

\begin{thebibliography}{100}
\providecommand{\url}[1]{#1}
\csname url@samestyle\endcsname
\providecommand{\newblock}{\relax}
\providecommand{\bibinfo}[2]{#2}
\providecommand{\BIBentrySTDinterwordspacing}{\spaceskip=0pt\relax}
\providecommand{\BIBentryALTinterwordstretchfactor}{4}
\providecommand{\BIBentryALTinterwordspacing}{\spaceskip=\fontdimen2\font plus
\BIBentryALTinterwordstretchfactor\fontdimen3\font minus
  \fontdimen4\font\relax}
\providecommand{\BIBforeignlanguage}[2]{{%
\expandafter\ifx\csname l@#1\endcsname\relax
\typeout{** WARNING: IEEEtran.bst: No hyphenation pattern has been}%
\typeout{** loaded for the language `#1'. Using the pattern for}%
\typeout{** the default language instead.}%
\else
\language=\csname l@#1\endcsname
\fi
#2}}
\providecommand{\BIBdecl}{\relax}
\BIBdecl

\bibitem{abu2012learning}
Y.~S. Abu-Mostafa, M.~Magdon-Ismail, and H.-T. Lin, \emph{Learning from
  data}.\hskip 1em plus 0.5em minus 0.4em\relax AMLBook New York, NY, USA,
  2012, vol.~4.

\bibitem{arulkumaran2017deep}
K.~Arulkumaran, M.~P. Deisenroth, M.~Brundage, and A.~A. Bharath, ``Deep
  reinforcement learning: A brief survey,'' \emph{IEEE Signal Processing
  Magazine}, vol.~34, no.~6, pp. 26--38, Nov. 2017.

\bibitem{gu2017deep}
S.~Gu, E.~Holly, T.~Lillicrap, and S.~Levine, ``Deep reinforcement learning for
  robotic manipulation with asynchronous off-policy updates,'' in \emph{IEEE
  international conference on robotics and automation (ICRA)}.\hskip 1em plus
  0.5em minus 0.4em\relax IEEE, May 2017, pp. 3389--3396.

\bibitem{peng2017deeploco}
X.~B. Peng, G.~Berseth, K.~Yin, and M.~Van De~Panne, ``Deeploco: Dynamic
  locomotion skills using hierarchical deep reinforcement learning,'' \emph{ACM
  Transactions on Graphics (TOG)}, vol.~36, no.~4, p.~41, Dec. 2017.

\bibitem{fayjie2018driverless}
A.~R. Fayjie, S.~Hossain, D.~Oualid, and D.-J. Lee, ``Driverless car:
  Autonomous driving using deep reinforcement learning in urban environment,''
  in \emph{15th International Conference on Ubiquitous Robots (UR)}.\hskip 1em
  plus 0.5em minus 0.4em\relax IEEE, Jun. 2018, pp. 896--901.

\bibitem{raghu2017continuous}
A.~Raghu, M.~Komorowski, L.~A. Celi, P.~Szolovits, and M.~Ghassemi,
  ``Continuous state-space models for optimal sepsis treatment-a deep
  reinforcement learning approach,'' \emph{Proceedings of Machine Learning for
  Healthcare}, Apr 2017.

\bibitem{deng2016deep}
Y.~Deng, F.~Bao, Y.~Kong, Z.~Ren, and Q.~Dai, ``Deep direct reinforcement
  learning for financial signal representation and trading,'' \emph{IEEE
  transactions on neural networks and learning systems}, vol.~28, no.~3, pp.
  653--664, Feb. 2016.

\bibitem{franccois2016deep}
V.~Fran{\c{c}}ois-Lavet, D.~Taralla, D.~Ernst, and R.~Fonteneau, ``Deep
  reinforcement learning solutions for energy microgrids management,'' in
  \emph{European Workshop on Reinforcement Learning (EWRL)}, Dec. 2016.

\bibitem{madu2017urban}
C.~N. Madu, C.-h. Kuei, and P.~Lee, ``Urban sustainability management: A deep
  learning perspective,'' \emph{Sustainable Cities and Society}, vol.~30, pp.
  1--17, 2017.

\bibitem{luong2018applications}
N.~C. Luong, D.~T. Hoang, S.~Gong, D.~Niyato, P.~Wang, Y.-C. Liang, and D.~I.
  Kim, ``Applications of deep reinforcement learning in communications and
  networking: A survey,'' \emph{IEEE Communications Surveys and Tutorials},
  vol.~21, no.~4, pp. 3133 -- 3174, May 2019.

\bibitem{mnih2015human}
V.~Mnih, K.~Kavukcuoglu, D.~Silver, A.~A. Rusu, J.~Veness, M.~G. Bellemare,
  A.~Graves, M.~Riedmiller, A.~K. Fidjeland, G.~Ostrovski \emph{et~al.},
  ``Human-level control through deep reinforcement learning,'' \emph{Nature},
  vol. 518, no. 7540, p. 529, Jun. 2015.

\bibitem{silver2016mastering}
D.~Silver, A.~Huang, C.~J. Maddison, A.~Guez, L.~Sifre, G.~Van Den~Driessche,
  J.~Schrittwieser, I.~Antonoglou, V.~Panneershelvam, M.~Lanctot \emph{et~al.},
  ``Mastering the game of {Go} with deep neural networks and tree search,''
  \emph{Nature}, vol. 529, no. 7587, p. 484, Dec. 2016.

\bibitem{silver2017mastering}
D.~Silver, T.~Hubert, J.~Schrittwieser, I.~Antonoglou, M.~Lai, A.~Guez,
  M.~Lanctot, L.~Sifre, D.~Kumaran, T.~Graepel \emph{et~al.}, ``A general
  reinforcement learning algorithm that masters chess, shogi, and go through
  self-play,'' \emph{Science}, vol. 362, no. 6419, pp. 1140--1144, 2018.

\bibitem{berner2019dota}
C.~Berner, G.~Brockman, B.~Chan, V.~Cheung, P.~D{\k{e}}biak, C.~Dennison,
  D.~Farhi, Q.~Fischer, S.~Hashme, C.~Hesse \emph{et~al.}, ``Dota 2 with large
  scale deep reinforcement learning,'' \emph{arXiv preprint arXiv:1912.06680},
  2019.

\bibitem{kiran2020deep}
B.~R. Kiran, I.~Sobh, V.~Talpaert, P.~Mannion, A.~A. Al~Sallab, S.~Yogamani,
  and P.~P{\'e}rez, ``Deep reinforcement learning for autonomous driving: A
  survey,'' \emph{IEEE Transactions on Intelligent Transportation Systems},
  2021.

\bibitem{zhang2020deep}
Z.~Zhang, S.~Zohren, and S.~Roberts, ``Deep reinforcement learning for
  trading,'' \emph{The Journal of Financial Data Science}, vol.~2, no.~2, pp.
  25--40, 2020.

\bibitem{behzadan2017vulnerability}
V.~Behzadan and A.~Munir, ``Vulnerability of deep reinforcement learning to
  policy induction attacks,'' in \emph{International Conference on Machine
  Learning and Data Mining in Pattern Recognition}.\hskip 1em plus 0.5em minus
  0.4em\relax Springer, Jul. 2017, pp. 262--275.

\bibitem{akhtar2018threat}
N.~Akhtar and A.~Mian, ``Threat of adversarial attacks on deep learning in
  computer vision: A survey,'' \emph{IEEE Access}, vol.~6, pp.
  14\,410--14\,430, 2018.

\bibitem{behzadan2018faults}
V.~Behzadan and A.~Munir, ``The faults in our pi stars: Security issues and
  open challenges in deep reinforcement learning,'' \emph{arXiv preprint
  arXiv:1810.10369}, Oct. 2018.

\bibitem{OpenAISpinUp1}
\BIBentryALTinterwordspacing
{OpenAI}, 2018, [Accessed 21 Jan. 2021.]. [Online]. Available:
  \url{https://spinningup.openai.com/en/latest/spinningup/rl_intro2.html}
\BIBentrySTDinterwordspacing

\bibitem{li2017deep}
Y.~Li, ``Deep reinforcement learning,'' \emph{arXiv preprint arXiv:1810.06339},
  Oct. 2018.

\bibitem{tramer2017space}
F.~Tram{\`e}r, N.~Papernot, I.~Goodfellow, D.~Boneh, and P.~McDaniel, ``The
  space of transferable adversarial examples,'' \emph{arXiv preprint
  arXiv:1704.03453}, May 2017.

\bibitem{cheng2018query}
M.~Cheng, T.~Le, P.-Y. Chen, J.~Yi, H.~Zhang, and C.-J. Hsieh,
  ``Query-efficient hard-label black-box attack: An optimization-based
  approach,'' \emph{arXiv preprint arXiv:1807.04457}, 2018.

\bibitem{chen2017zoo}
P.-Y. Chen, H.~Zhang, Y.~Sharma, J.~Yi, and C.-J. Hsieh, ``{ZOO}: Zeroth order
  optimization based black-box attacks to deep neural networks without training
  substitute models,'' in \emph{Proceedings of the 10th ACM Workshop on
  Artificial Intelligence and Security}, 2017, pp. 15--26.

\bibitem{tu2019autozoom}
C.-C. Tu, P.~Ting, P.-Y. Chen, S.~Liu, H.~Zhang, J.~Yi, C.-J. Hsieh, and S.-M.
  Cheng, ``Autozoom: Autoencoder-based zeroth order optimization method for
  attacking black-box neural networks,'' in \emph{Proceedings of the AAAI
  Conference on Artificial Intelligence}, vol.~33, 2019, pp. 742--749.

\bibitem{zhang2019adversarial}
J.~Zhang and C.~Li, ``Adversarial examples: Opportunities and challenges,''
  \emph{IEEE transactions on neural networks and learning systems}, 2019.

\bibitem{qayyum2019securing}
A.~Qayyum, M.~Usama, J.~Qadir, and A.~Al-Fuqaha, ``Securing connected \&
  autonomous vehicles: Challenges posed by adversarial machine learning and the
  way forward,'' \emph{IEEE Communications Surveys \& Tutorials}, vol.~22,
  no.~2, pp. 998--1026, 2020.

\bibitem{clark2018malicious}
G.~Clark, M.~Doran, and W.~Glisson, ``A malicious attack on the machine
  learning policy of a robotic system,'' in \emph{2018 17th IEEE International
  Conference On Trust, Security And Privacy In Computing And
  Communications/12th IEEE International Conference On Big Data Science And
  Engineering (TrustCom/BigDataSE)}.\hskip 1em plus 0.5em minus 0.4em\relax
  IEEE, Aug. 2018, pp. 516--521.

\bibitem{vorobeychik2018adversarial}
Y.~Vorobeychik and M.~Kantarcioglu, ``Adversarial machine learning,''
  \emph{Synthesis Lectures on Artificial Intelligence and Machine Learning},
  vol.~12, no.~3, pp. 1--169, Jun. 2018.

\bibitem{papernot2016limitations}
N.~Papernot, P.~McDaniel, S.~Jha, M.~Fredrikson, Z.~B. Celik, and A.~Swami,
  ``The limitations of deep learning in adversarial settings,'' in
  \emph{Security and Privacy (EuroS\&P), 2016 IEEE European Symposium
  on}.\hskip 1em plus 0.5em minus 0.4em\relax IEEE, Mar. 2016, pp. 372--387.

\bibitem{huang2017adversarial}
S.~Huang, N.~Papernot, I.~Goodfellow, Y.~Duan, and P.~Abbeel, ``Adversarial
  attacks on neural network policies,'' \emph{ICLR Workshop}, 3 2017.

\bibitem{lin2017tactics}
Y.-C. Lin, Z.-W. Hong, Y.-H. Liao, M.-L. Shih, M.-Y. Liu, and M.~Sun, ``Tactics
  of adversarial attack on deep reinforcement learning agents,''
  \emph{International Joint Conferences on Artificial Intelligence}, 8 2017.

\bibitem{carlini2017towards}
N.~Carlini and D.~Wagner, ``Towards evaluating the robustness of neural
  networks,'' in \emph{Security and Privacy (SP), 2017 IEEE Symposium
  on}.\hskip 1em plus 0.5em minus 0.4em\relax IEEE, May 2017, pp. 39--57.

\bibitem{tretschk2018sequential}
E.~Tretschk, S.~J. Oh, and M.~Fritz, ``Sequential attacks on agents for
  long-term adversarial goals,'' \emph{arXiv preprint arXiv:1805.12487}, Jul.
  2018.

\bibitem{baluja2018learning}
S.~Baluja and I.~Fischer, ``Learning to attack: Adversarial transformation
  networks,'' in \emph{Association for the Advancement of Artificial
  Intelligence}, Feb. 2018, pp. 2687--2695.

\bibitem{pattanaik2018robust}
A.~Pattanaik, Z.~Tang, S.~Liu, G.~Bommannan, and G.~Chowdhary, ``Robust deep
  reinforcement learning with adversarial attacks,'' in \emph{Proceedings of
  the 17th International Conference on Autonomous Agents and MultiAgent
  Systems}.\hskip 1em plus 0.5em minus 0.4em\relax International Foundation for
  Autonomous Agents and Multiagent Systems, Jul. 2018, pp. 2040--2042.

\bibitem{kos2017delving}
J.~Kos and D.~Song, ``Delving into adversarial attacks on deep policies,''
  \emph{ICLR Workshop}, Apr. 2017.

\bibitem{sun2020stealthy}
J.~Sun, T.~Zhang, X.~Xie, L.~Ma, Y.~Zheng, K.~Chen, and Y.~Liu, ``Stealthy and
  efficient adversarial attacks against deep reinforcement learning,'' in
  \emph{Proceedings of the AAAI Conference on Artificial Intelligence},
  vol.~34, no.~04, 2020, pp. 5883--5891.

\bibitem{hussenot2019targeted}
L.~Hussenot, M.~Geist, and O.~Pietquin, ``Targeted attacks on deep
  reinforcement learning agents through adversarial observations,''
  \emph{Autonomous Agents and Multi-Agent Systems ({AAMAS})}, May 2020.

\bibitem{chan2020adversarial}
P.~P. Chan, Y.~Wang, and D.~S. Yeung, ``Adversarial attack against deep
  reinforcement learning with static reward impact map,'' in \emph{Proceedings
  of the 15th ACM Asia Conference on Computer and Communications Security},
  2020, pp. 334--343.

\bibitem{de2018cooperative}
C.~de~Vrieze, S.~Barratt, D.~Tsai, and A.~Sahai, ``Cooperative multi-agent
  reinforcement learning for low-level wireless communication,'' \emph{arXiv
  preprint arXiv:1801.04541}, 2018.

\bibitem{wiering2000multi}
M.~Wiering, ``Multi-agent reinforcement learning for traffic light control,''
  in \emph{Machine Learning: Proceedings of the Seventeenth International
  Conference (ICML'2000)}, 2000, pp. 1151--1158.

\bibitem{shalev2016safe}
S.~Shalev-Shwartz, S.~Shammah, and A.~Shashua, ``Safe, multi-agent,
  reinforcement learning for autonomous driving,'' \emph{arXiv preprint
  arXiv:1610.03295}, Oct. 2016.

\bibitem{lin2020robustness}
J.~Lin, K.~Dzeparoska, S.~Q. Zhang, A.~Leon-Garcia, and N.~Papernot, ``On the
  robustness of cooperative multi-agent reinforcement learning,'' \emph{arXiv
  preprint arXiv:2003.03722}, 2020.

\bibitem{wang2020adversarial}
P.~Wang, Y.~Li, S.~Shekhar, and W.~F. Northrop, ``Adversarial attacks on
  reinforcement learning based energy management systems of extended range
  electric delivery vehicles,'' \emph{arXiv preprint arXiv:2006.00817}, 2020.

\bibitem{xiao2019characterizing}
C.~Xiao, X.~Pan, W.~He, J.~Peng, M.~Sun, J.~Yi, B.~Li, and D.~Song,
  ``Characterizing attacks on deep reinforcement learning,'' \emph{arXiv
  preprint arXiv:1907.09470}, Jul. 2019.

\bibitem{chen2018gradient}
T.~Chen, W.~Niu, Y.~Xiang, X.~Bai, J.~Liu, Z.~Han, and G.~Li, ``Gradient
  band-based adversarial training for generalized attack immunity of {A3C} path
  finding,'' \emph{arXiv preprint arXiv:1807.06752}, Jul. 2018.

\bibitem{bai2018adversarial}
X.~Bai, W.~Niu, J.~Liu, X.~Gao, Y.~Xiang, and J.~Liu, ``Adversarial examples
  construction towards white-box {Q}-table variation in {DQN} pathfinding
  training,'' in \emph{IEEE Third International Conference on Data Science in
  Cyberspace (DSC)}.\hskip 1em plus 0.5em minus 0.4em\relax IEEE, Jun. 2018,
  pp. 781--787.

\bibitem{gleave2019adversarial}
\BIBentryALTinterwordspacing
A.~Gleave, M.~Dennis, C.~Wild, N.~Kant, S.~Levine, and S.~Russell,
  ``Adversarial policies: Attacking deep reinforcement learning,'' in
  \emph{International Conference on Learning Representations}, 2020. [Online].
  Available: \url{https://openreview.net/forum?id=HJgEMpVFwB}
\BIBentrySTDinterwordspacing

\bibitem{yang2020enhanced}
C.-H.~H. Yang, J.~Qi, P.-Y. Chen, Y.~Ouyang, I.-T.~D. Hung, C.-H. Lee, and
  X.~Ma, ``Enhanced adversarial strategically-timed attacks against deep
  reinforcement learning,'' in \emph{ICASSP 2020-2020 IEEE International
  Conference on Acoustics, Speech and Signal Processing (ICASSP)}.\hskip 1em
  plus 0.5em minus 0.4em\relax IEEE, 2020, pp. 3407--3411.

\bibitem{kiourti2019trojdrl}
K.~Panagiota, W.~Kacper, S.~Jha, and L.~Wenchao, ``{TrojDRL}: Trojan attacks on
  deep reinforcement learning agents.'' in \emph{Proc. 57th ACM/IEEE Design
  Automation Conference (DAC), 2020}, Mar. 2020.

\bibitem{behzadan2019sequential}
V.~Behzadan and W.~Hsu, ``Sequential triggers for watermarking of deep
  reinforcement learning policies,'' \emph{arXiv preprint arXiv:1906.01126},
  Jun. 2019.

\bibitem{usama2020examining}
M.~Usama, R.~Mitra, I.~Ilahi, J.~Qadir, and M.~Marina, ``Examining machine
  learning for {5G} and beyond through an adversarial lens,'' \emph{IEEE
  Internet Computing}, 1 2021.

\bibitem{han2018reinforcement}
Y.~Han, B.~I. Rubinstein, T.~Abraham, T.~Alpcan, O.~De~Vel, S.~Erfani,
  D.~Hubczenko, C.~Leckie, and P.~Montague, ``Reinforcement learning for
  autonomous defence in software-defined networking,'' in \emph{International
  Conference on Decision and Game Theory for Security}.\hskip 1em plus 0.5em
  minus 0.4em\relax Springer, Aug. 2018, pp. 145--165.

\bibitem{huang2019deceptive}
Y.~Huang and Q.~Zhu, ``Deceptive reinforcement learning under adversarial
  manipulations on cost signals,'' in \emph{International Conference on
  Decision and Game Theory for Security}.\hskip 1em plus 0.5em minus
  0.4em\relax Springer, 2019, pp. 217--237.

\bibitem{rakhsha2020policy}
A.~Rakhsha, G.~Radanovic, R.~Devidze, X.~Zhu, and A.~Singla, ``Policy teaching
  via environment poisoning: Training-time adversarial attacks against
  reinforcement learning,'' in \emph{International Conference on Machine
  Learning}.\hskip 1em plus 0.5em minus 0.4em\relax PMLR, 2020, pp. 7974--7984.

\bibitem{yeow2019spatiotemporally}
X.~Y. Lee, S.~Ghadai, K.~L. Tan, C.~Hegde, and S.~Sarkar, ``Spatiotemporally
  constrained action space attacks on deep reinforcement learning agents,'' in
  \emph{Proceedings of the AAAI Conference on Artificial Intelligence},
  vol.~34, no.~04, 2020, pp. 4577--4584.

\bibitem{behzadan2019adversarial}
V.~Behzadan and W.~Hsu, ``Adversarial exploitation of policy imitation,''
  \emph{arXiv preprint arXiv:1906.01121}, Jun. 2019.

\bibitem{hessel2018rainbow}
M.~Hessel, J.~Modayil, H.~Van~Hasselt, T.~Schaul, G.~Ostrovski, W.~Dabney,
  D.~Horgan, B.~Piot, M.~Azar, and D.~Silver, ``Rainbow: Combining improvements
  in deep reinforcement learning,'' in \emph{Thirty-Second AAAI Conference on
  Artificial Intelligence}, Feb. 2018.

\bibitem{pan2017virtual}
X.~Pan, Y.~You, Z.~Wang, and C.~Lu, ``Virtual to real reinforcement learning
  for autonomous driving,'' \emph{Proceedings of the British Machine Vision
  Conference (BMVC)}, Sep. 2017.

\bibitem{chen2020stealing}
K.~Chen, S.~Guo, T.~Zhang, X.~Xie, and Y.~Liu, ``Stealing deep reinforcement
  learning models for fun and profit,'' in \emph{Proceedings of the 2021 ACM
  Asia Conference on Computer and Communications Security}, 2021, pp. 307--319.

\bibitem{wang2016sample}
Z.~Wang, V.~Bapst, N.~Heess, V.~Mnih, R.~Munos, K.~Kavukcuoglu, and
  N.~de~Freitas, ``Sample efficient actor-critic with experience replay,''
  \emph{ICLR}, 2017.

\bibitem{wu2017scalable}
Y.~Wu, E.~Mansimov, R.~B. Grosse, S.~Liao, and J.~Ba, ``Scalable trust-region
  method for deep reinforcement learning using kronecker-factored
  approximation,'' in \emph{Advances in neural information processing systems},
  2017, pp. 5279--5288.

\bibitem{huai2020malicious}
M.~Huai, J.~Sun, R.~Cai, L.~Yao, and A.~Zhang, ``Malicious attacks against deep
  reinforcement learning interpretations,'' in \emph{Proceedings of the 26th
  ACM SIGKDD International Conference on Knowledge Discovery \& Data Mining},
  2020, pp. 472--482.

\bibitem{lee2020query}
X.~Y. Lee, Y.~Esfandiari, K.~L. Tan, and S.~Sarkar, ``Query-based targeted
  action-space adversarial policies on deep reinforcement learning agents,'' in
  \emph{Proceedings of the ACM/IEEE 12th International Conference on
  Cyber-Physical Systems}, 2021, pp. 87--97.

\bibitem{ray2019benchmarking}
A.~Ray, J.~Achiam, and D.~Amodei, ``Benchmarking safe exploration in deep
  reinforcement learning,'' \emph{arXiv preprint arXiv:1910.01708}, 2019.

\bibitem{behzadan2017whatever}
\BIBentryALTinterwordspacing
V.~Behzadan and A.~Munir, ``Whatever does not kill deep reinforcement learning,
  makes it stronger,'' \emph{arXiv preprint arXiv:1712.09344}, Dec. 2017.
  [Online]. Available: \url{https://github.com/behzadanksu/rl-attack}
\BIBentrySTDinterwordspacing

\bibitem{behzadan2018mitigation}
------, ``Mitigation of policy manipulation attacks on deep q-networks with
  parameter-space noise,'' in \emph{International Conference on Computer
  Safety, Reliability, and Security}.\hskip 1em plus 0.5em minus 0.4em\relax
  Springer, 2018, pp. 406--417.

\bibitem{fortunato2017noisy}
M.~Fortunato, M.~G. Azar, B.~Piot, J.~Menick, I.~Osband, A.~Graves, V.~Mnih,
  R.~Munos, D.~Hassabis, O.~Pietquin \emph{et~al.}, ``Noisy networks for
  exploration,'' \emph{ICLR}, May 2018.

\bibitem{behzadan2019analysis}
V.~Behzadan and W.~Hsu, ``Analysis and improvement of adversarial training in
  {DQN} agents with adversarially-guided exploration ({AGE}),'' \emph{arXiv
  preprint arXiv:1906.01119}, Jun. 2019.

\bibitem{tan2020robustifying}
K.~L. Tan, Y.~Esfandiari, X.~Y. Lee, S.~Sarkar \emph{et~al.}, ``Robustifying
  reinforcement learning agents via action space adversarial training,'' in
  \emph{2020 American control conference (ACC)}.\hskip 1em plus 0.5em minus
  0.4em\relax IEEE, 2020, pp. 3959--3964.

\bibitem{vinitsky2020robust}
E.~Vinitsky, Y.~Du, K.~Parvate, K.~Jang, P.~Abbeel, and A.~Bayen, ``Robust
  reinforcement learning using adversarial populations,'' \emph{arXiv preprint
  arXiv:2008.01825}, 2020.

\bibitem{pinto2017robust}
L.~Pinto, J.~Davidson, R.~Sukthankar, and A.~Gupta, ``Robust adversarial
  reinforcement learning,'' in \emph{Proceedings of the 34th International
  Conference on Machine Learning-Volume 70}.\hskip 1em plus 0.5em minus
  0.4em\relax JMLR. org, Aug. 2017, pp. 2817--2826.

\bibitem{abdullah2019wasserstein}
M.~A. Abdullah, H.~Ren, H.~B. Ammar, V.~Milenkovic, R.~Luo, M.~Zhang, and
  J.~Wang, ``Wasserstein robust reinforcement learning,'' \emph{arXiv preprint
  arXiv:1907.13196}, Sep. 2019.

\bibitem{bravo2017robustness}
M.~Bravo and P.~Mertikopoulos, ``On the robustness of learning in games with
  stochastically perturbed payoff observations,'' \emph{Games and Economic
  Behavior}, vol. 103, pp. 41--66, May 2017.

\bibitem{ogunmolu2018minimax}
O.~Ogunmolu, N.~Gans, and T.~Summers, ``Minimax iterative dynamic game:
  Application to nonlinear robot control tasks,'' in \emph{2018 IEEE/RSJ
  International Conference on Intelligent Robots and Systems (IROS)}.\hskip 1em
  plus 0.5em minus 0.4em\relax IEEE, Oct. 2018, pp. 6919--6925.

\bibitem{mandlekar2017adversarially}
A.~Mandlekar, Y.~Zhu, A.~Garg, L.~Fei-Fei, and S.~Savarese, ``Adversarially
  robust policy learning: Active construction of physically-plausible
  perturbations,'' in \emph{2017 IEEE/RSJ International Conference on
  Intelligent Robots and Systems (IROS)}.\hskip 1em plus 0.5em minus
  0.4em\relax IEEE, Sep. 2017, pp. 3932--3939.

\bibitem{wang2018reinforcement}
J.~Wang, Y.~Liu, and B.~Li, ``Reinforcement learning with perturbed rewards,''
  \emph{Thirty-forth {AAAI} Conference on Artificial Intelligence}, Feb. 2020.

\bibitem{smirnova2019distributionally}
E.~Smirnova, E.~Dohmatob, and J.~Mary, ``Distributionally robust reinforcement
  learning,'' \emph{ICML Workshop}, May 2019.

\bibitem{tessler2019action}
C.~Tessler, Y.~Efroni, and S.~Mannor, ``Action robust reinforcement learning
  and applications in continuous control,'' \emph{ICML}, Jun. 2019.

\bibitem{kumar2019enhancing}
A.~Kumar \emph{et~al.}, ``Enhancing performance of reinforcement learning
  models in the presence of noisy rewards,'' Ph.D. dissertation, University of
  Texas, Austin, 2019.

\bibitem{fischer2019online}
M.~Fischer, M.~Mirman, and M.~Vechev, ``Online robustness training for deep
  reinforcement learning,'' \emph{arXiv preprint arXiv:1911.00887}, Nov. 2019.

\bibitem{pan2019risk}
X.~Pan, D.~Seita, Y.~Gao, and J.~Canny, ``Risk averse robust adversarial
  reinforcement learning,'' in \emph{2019 International Conference on Robotics
  and Automation (ICRA)}.\hskip 1em plus 0.5em minus 0.4em\relax IEEE, 2019,
  pp. 8522--8528.

\bibitem{osband2016deep}
I.~Osband, C.~Blundell, A.~Pritzel, and B.~Van~Roy, ``Deep exploration via
  bootstrapped {DQN},'' in \emph{Advances in neural information processing
  systems}, 2016, pp. 4026--4034.

\bibitem{wang2020falsification}
X.~Wang, S.~Nair, and M.~Althoff, ``Falsification-based robust adversarial
  reinforcement learning,'' in \emph{2020 19th IEEE International Conference on
  Machine Learning and Applications (ICMLA)}.\hskip 1em plus 0.5em minus
  0.4em\relax IEEE, 2020, pp. 205--212.

\bibitem{lutjens2020certified}
B.~L{\"u}tjens, M.~Everett, and J.~P. How, ``Certified adversarial robustness
  for deep reinforcement learning,'' in \emph{Conference on Robot Learning},
  2020, pp. 1328--1337.

\bibitem{zhang2020robust}
\BIBentryALTinterwordspacing
H.~Zhang, H.~Chen, C.~Xiao, B.~Li, M.~Liu, D.~Boning, and C.-J. Hsieh, ``Robust
  deep reinforcement learning against adversarial perturbations on state
  observations,'' in \emph{Advances in Neural Information Processing Systems},
  H.~Larochelle, M.~Ranzato, R.~Hadsell, M.~F. Balcan, and H.~Lin, Eds.,
  vol.~33.\hskip 1em plus 0.5em minus 0.4em\relax Curran Associates, Inc.,
  2020, pp. 21\,024--21\,037. [Online]. Available:
  \url{https://proceedings.neurips.cc/paper/2020/file/f0eb6568ea114ba6e293f903c34d7488-Paper.pdf}
\BIBentrySTDinterwordspacing

\bibitem{oikarinen2020robust}
T.~Oikarinen, T.-W. Weng, and L.~Daniel, ``Robust deep reinforcement learning
  through adversarial loss,'' \emph{arXiv preprint arXiv:2008.01976}, 2020.

\bibitem{zhang2021robust}
H.~Zhang, H.~Chen, D.~Boning, and C.-J. Hsieh, ``Robust reinforcement learning
  on state observations with learned optimal adversary,'' \emph{arXiv preprint
  arXiv:2101.08452}, 2021.

\bibitem{lin2017detecting}
Y.-C. Lin, M.-Y. Liu, M.~Sun, and J.-B. Huang, ``Detecting adversarial attacks
  on neural network policies with visual foresight,'' \emph{arXiv preprint
  arXiv:1710.00814}, Oct. 2017.

\bibitem{kurakin2016adversarial}
A.~Kurakin, I.~Goodfellow, and S.~Bengio, ``Adversarial examples in the
  physical world,'' \emph{arXiv preprint arXiv:1607.02533}, Feb. 2017.

\bibitem{havens2018online}
A.~Havens, Z.~Jiang, and S.~Sarkar, ``Online robust policy learning in the
  presence of unknown adversaries,'' in \emph{Advances in Neural Information
  Processing Systems}, Dec. 2018, pp. 9916--9926.

\bibitem{xiang2018pca}
Y.~Xiang, W.~Niu, J.~Liu, T.~Chen, and Z.~Han, ``A {PCA}-based model to predict
  adversarial examples on {Q}-learning of path finding,'' in \emph{2018 IEEE
  Third International Conference on Data Science in Cyberspace (DSC)}.\hskip
  1em plus 0.5em minus 0.4em\relax IEEE, Jun. 2018, pp. 773--780.

\bibitem{gallego2019reinforcement}
V.~Gallego, R.~Naveiro, and D.~R. Insua, ``Reinforcement learning under
  threats,'' in \emph{Proceedings of the AAAI Conference on Artificial
  Intelligence}, vol.~33, Jul. 2019, pp. 9939--9940.

\bibitem{papernot2016distillation}
N.~Papernot, P.~McDaniel, X.~Wu, S.~Jha, and A.~Swami, ``Distillation as a
  defense to adversarial perturbations against deep neural networks,'' in
  \emph{2016 IEEE Symposium on Security and Privacy (SP)}.\hskip 1em plus 0.5em
  minus 0.4em\relax IEEE, May 2016, pp. 582--597.

\bibitem{carlini2016defensive}
N.~Carlini and D.~Wagner, ``Defensive distillation is not robust to adversarial
  examples,'' \emph{arXiv preprint arXiv:1607.04311}, Jul. 2016.

\bibitem{rusu2015policy}
A.~A. Rusu, S.~G. Colmenarejo, C.~Gulcehre, G.~Desjardins, J.~Kirkpatrick,
  R.~Pascanu, V.~Mnih, K.~Kavukcuoglu, and R.~Hadsell, ``Policy distillation,''
  \emph{ICLR}, May 2016.

\bibitem{pmlr-v89-czarnecki19a}
W.~M. Czarnecki, R.~Pascanu, S.~Osindero, S.~Jayakumar, G.~Swirszcz, and
  M.~Jaderberg, ``Distilling policy distillation,'' in \emph{Proceedings of
  Machine Learning Research}, Apr 2019, pp. 1331--1340.

\bibitem{qu2020defending}
X.~Qu, Y.-S. Ong, A.~Gupta, and Z.~Sun, ``Defending adversarial attacks without
  adversarial attacks in deep reinforcement learning,'' \emph{arXiv preprint
  arXiv:2008.06199}, 2020.

\bibitem{behzadan2018adversarial}
V.~Behzadan and A.~Munir, ``Adversarial reinforcement learning framework for
  benchmarking collision avoidance mechanisms in autonomous vehicles,''
  \emph{IEEE Intelligent Transportation Systems Magazine}, Apr 2019.

\bibitem{behzadan2019rl}
V.~Behzadan and W.~Hsu, ``{RL}-based method for benchmarking the adversarial
  resilience and robustness of deep reinforcement learning policies,'' in
  \emph{International Conference on Computer Safety, Reliability, and
  Security}.\hskip 1em plus 0.5em minus 0.4em\relax Springer, 2019, pp.
  314--325.

\bibitem{1606.01540}
G.~Brockman, V.~Cheung, L.~Pettersson, J.~Schneider, J.~Schulman, J.~Tang, and
  W.~Zaremba, ``{OpenAI Gym},'' \emph{arXiv preprint arXiv:1606.01540}, 2016.

\bibitem{tensorflow2015-whitepaper}
\BIBentryALTinterwordspacing
Abadi \emph{et~al.}, ``{TensorFlow}: Large-scale machine learning on
  heterogeneous systems,'' 2015, software available from tensorflow.org.
  [Online]. Available: \url{https://www.tensorflow.org/}
\BIBentrySTDinterwordspacing

\bibitem{baselines}
P.~Dhariwal, C.~Hesse, O.~Klimov, A.~Nichol, M.~Plappert, A.~Radford,
  J.~Schulman, S.~Sidor, Y.~Wu, and P.~Zhokhov, ``{OpenAI} baselines,''
  \url{https://github.com/openai/baselines}, 2017.

\bibitem{caspi_itai_2017_1134899}
\BIBentryALTinterwordspacing
I.~Caspi, G.~Leibovich, G.~Novik, and S.~Endrawis, ``Reinforcement learning
  coach,'' Dec 2017. [Online]. Available:
  \url{https://doi.org/10.5281/zenodo.1134899}
\BIBentrySTDinterwordspacing

\bibitem{gauci2018horizon}
J.~Gauci, E.~Conti, Y.~Liang, K.~Virochsiri, Y.~He, Z.~Kaden, V.~Narayanan, and
  X.~Ye, ``Horizon: Facebook's open source applied reinforcement learning
  platform,'' \emph{arXiv preprint arXiv:1811.00260}, Sep. 2019.

\bibitem{reagent}
\BIBentryALTinterwordspacing
{ReAgent}, ``{ReAgent”},'' [Accessed Jan. 21, 2021.]. [Online]. Available:
  \url{https://reagent.ai/}
\BIBentrySTDinterwordspacing

\bibitem{ns3gym}
\BIBentryALTinterwordspacing
P.~Gaw{\l}owicz and A.~Zubow, ``{{ns}-3 meets {OpenAI} Gym: The Playground for
  Machine Learning in Networking Research},'' in \emph{{ACM International
  Conference on Modeling, Analysis and Simulation of Wireless and Mobile
  Systems (MSWiM)}}, 11 2019. [Online]. Available:
  \url{http://www.tkn.tu-berlin.de/fileadmin/fg112/Papers/2019/gawlowicz19_mswim.pdf}
\BIBentrySTDinterwordspacing

\bibitem{papernot2018cleverhans}
Papernot \emph{et~al.}, ``Technical report on the {C}leverhans v2.1.0
  adversarial examples library,'' \emph{arXiv preprint arXiv: 1610.00768},
  2018.

\bibitem{moosavi2017analysis}
S.-M. Moosavi-Dezfooli, A.~Fawzi, O.~Fawzi, and P.~Frossard, ``Universal
  adversarial perturbations,'' in \emph{Proceedings of the IEEE Conference on
  Computer Vision and Pattern Recognition (CVPR)}, July 2017.

\bibitem{puiutta2020explainable}
E.~Puiutta and E.~M. Veith, ``Explainable reinforcement learning: A survey,''
  in \emph{International Cross-Domain Conference for Machine Learning and
  Knowledge Extraction}.\hskip 1em plus 0.5em minus 0.4em\relax Springer, 2020,
  pp. 77--95.

\bibitem{heuillet2021explainability}
A.~Heuillet, F.~Couthouis, and N.~D{\'\i}az-Rodr{\'\i}guez, ``Explainability in
  deep reinforcement learning,'' \emph{Knowledge-Based Systems}, vol. 214, p.
  106685, 2021.

\bibitem{tan2018survey}
C.~Tan, F.~Sun, T.~Kong, W.~Zhang, C.~Yang, and C.~Liu, ``A survey on deep
  transfer learning,'' in \emph{International conference on artificial neural
  networks}.\hskip 1em plus 0.5em minus 0.4em\relax Springer, 2018, pp.
  270--279.

\bibitem{zhu2020transfer}
Z.~Zhu, K.~Lin, and J.~Zhou, ``Transfer learning in deep reinforcement
  learning: A survey,'' \emph{arXiv preprint arXiv:2009.07888}, 2020.

\end{thebibliography}

\end{document}